%% file: preprint.tex
\documentclass[sigconf]{acmart}
\AtBeginDocument{%
  \providecommand\BibTeX{{%
    Bib\TeX}}}
\usepackage{multirow}
\usepackage{hyperref}
\usepackage[nameinlink,capitalise]{cleveref}

\usepackage{enumitem}
\usepackage[table,dvipsnames]{xcolor}
\setlist[enumerate]{noitemsep, topsep=0pt, parsep=0pt, partopsep=0pt}
\usepackage{graphicx}
\usepackage{float}
\usepackage{xcolor}
\usepackage{subcaption}
\usepackage{float}
\usepackage{algorithm}
\usepackage{algorithmicx}
\usepackage{algpseudocode}
\usepackage{duckuments}
\usepackage{pgfplots}
\definecolor{mycitecolor}{RGB}{0, 102, 204}  

\usepackage{listings}
\definecolor{codekeyword}{rgb}{0.74, 0.0, 0.44}   
\definecolor{codecomment}{rgb}{0.0, 0.5, 0.5}     
\lstdefinestyle{pytorch}{
    commentstyle=\color{codecomment},
    keywordstyle=\color{codekeyword}\bfseries,
    basicstyle=\ttfamily\small,
    breaklines=true,
    columns=fullflexible,
    keepspaces=true,
    showstringspaces=false,
    language=Python,
    morekeywords={def, for, in, range, return},
    frame=none,
    xleftmargin=0pt,
    xrightmargin=0pt,
}
\usepackage{pifont}

\newcommand{\xmark}{\textcolor{red}{\ding{55}}}
\newcommand{\checkmarkb}{\textcolor{green!60!black}{\ding{52}}}

\newcommand{\Nalpha}{\ensuremath{N_\alpha}}

\usepackage{bm}
\makeatletter
\hypersetup{
  colorlinks=true,
  citecolor=mycitecolor,
  linkcolor=black,
  urlcolor=blue
}
\makeatother
\begin{document}

\title{MAPLE: Efficient and Diverse Multi-Alpha Generation for Portfolio Construction}


\definecolor{darkgreen}{rgb}{0.00, 0.40, 0.00}


\newcommand{\abner}[1]{\textcolor{blue}{[@abner: #1]}}
\newcommand{\lava}[1]{\textcolor{purple}{[@lava: #1]}}
\newcommand{\model}[1]{{MAPLE}}

\author{Yu-Chen Den}\authornote{Equal contribution}
\affiliation{%
  \institution{SinoPac Holdings}
  \city{Taipei}
  \country{Taiwan}
}
\email{abnerden@sinopac.com}
\authornote{Corresponding author}

\author{Kuan-Yu Chen}\authornotemark[1]
\affiliation{%
  \institution{SinoPac Holdings}
  \city{Taipei}
  \country{Taiwan}
}
\email{lavamore@sinopac.com}

\author{Kendro Vincent}
\affiliation{%
    \institution{National Chengchi University}
    \city{Taipei}
    \country{Taiwan}
}
\email{kendrov@g.nccu.edu.tw}

\author{Tien-Hao Chang}
\affiliation{%
  \institution{SinoPac Holdings}
  \city{Taipei}
  \country{Taiwan}
}
\email{darby@sinopac.com}

\renewcommand{\shortauthors}{Den and Chen et al.}

\begin{abstract}
Classical alpha mining achieves strong risk-adjusted returns by combining many low-correlated predictive signals, yet deep learning stock-ranking methods typically produce a single alpha per stock, rely on increasingly complex architectures with diminishing gains, and obtain diversity only through separate models or implicit routing, without explicitly controlling inter-alpha correlation. We introduce \model{} (\textbf{M}ulti-\textbf{A}lpha \textbf{P}osition-aware \textbf{L}istwise \textbf{E}nsembling), a backbone-agnostic framework that recovers this diversity principle within a single training pass. \model{} combines a unified, capacity-scaled prediction head with an extreme-rank weighted listwise ranking loss and a diversity regularizer that explicitly penalizes pairwise correlation across alphas. Across four equity markets spanning the US, China, and Japan, \model{} achieves the best average Sharpe and Calmar ratios among nine baselines, 
using up to 55$\times$ fewer parameters and 2.5$\times$ less training time, 
and generalizes across five backbone architectures with Sharpe and Calmar Ratio gains of 10--23\% and 17--43\%, respectively. Behavioral analysis further shows why each component works: the unified head already reduces inter-alpha correlation before any diversity loss is applied, and the extreme-rank loss lets diversity regularization improve rather than erode per-alpha ranking quality as capacity scaling sustains this balance at scale. These results show that principled loss design and capacity allocation, rather than architectural complexity, drive diverse and effective multi-alpha generation.
\end{abstract}

\ccsdesc[500]{Information systems~Data mining}
\ccsdesc[500]{Applied computing~Economics}
\ccsdesc[500]{Computing methodologies~Machine learning}

\keywords{Quantitative Finance, Alpha Mining, Ensemble, Learning to Rank, Self Attention}



\maketitle

\input{content/intro}
\input{content/prelim-new}
\input{content/method}
\input{content/expr}
\input{content/analysis}

\input{content/related}
\input{content/conclude}

\bibliographystyle{ACM-Reference-Format}
\bibliography{neural_alpha}

\clearpage
\appendix
\input{content/appendix}
\end{document}

%% file: content/intro.tex
\section{Introduction}
\label{sec:intro}

Classical alpha mining aims to discover predictive signals that can be transformed into trading strategies and ultimately combined into portfolios. Traditionally, such signals relied on economic intuition, fundamental analysis, and handcrafted features~\citep{fama2018choosing}. To overcome the pace limitation of manual discovery, researchers have explored automated approaches including genetic programming~\citep{lin2019genetic}, reinforcement learning~\citep{zhang2020autoalpha,shi2025alphaforge,yu2023generating}, and LLM-based agents~\citep{tang2025alphaagent,shi2026navigating}. Among these, deep learning stock ranking has emerged as the dominant paradigm for alpha signal learning: it is end-to-end trainable directly from price and volume data, scales naturally with data availability, and has produced strong empirical results across diverse market settings~\citep{fang2019alpha,hsu2021fingat,li2024master,sawhney2021stock}. The core workflow is straightforward---a model predicts the cross-sectional return of each stock, and a portfolio is constructed by selecting the top-ranked names.

Despite this progress, two fundamental limitations remain unaddressed. First, in pursuit of better single-model performance, recent architectures have grown increasingly complex---stacking multi-layer Transformers with intra- and inter-stock attention~\citep{li2024master}, or incorporating graph networks~\citep{hsu2021fingat}, Mixture-of-Experts routing~\citep{yu2024miga,liu2025mera,chen2025dhmoe}, and diffusion-based components~\citep{chen2025dhmoe}. While these designs improve expressiveness, they yield diminishing performance gains at substantially higher computational cost. Simpler architectures have been shown to match or outperform complex Transformer variants on financial tasks~\citep{den2026integrating}, suggesting that architectural complexity alone is not the bottleneck. Moreover, their architecture-specific nature makes them difficult to transfer or extend across different backbone settings, as each new design requires substantial re-engineering to apply elsewhere.
 
Second, a central principle in classical alpha mining is signal diversity: individual alphas are susceptible to regime shifts and structural changes, making low-correlated ensembles essential for managing concentrated exposure and improving risk-adjusted returns~\citep{goldstein2015information,acharya2017measuring,tulchinsky2019finding}. Yet most deep learning methods generate only a single alpha score per stock~\citep{yoo2021accurate,li2024master,sawhney2021stock,xia2024cisthpan,liu2025mera}, overlooking the diversification benefit that multi-alpha ensembles provide. Attempts to address this have relied on training multiple independent models, either with passive diversity through random seeds or architectural variation~\citep{sun2023mastering,qin2026fineft}, with explicit routing mechanisms~\citep{yu2024miga,liu2025mera,chen2025dhmoe}, or with multi-stage distillation training~\citep{den2026integrating}.
All of these methods multiply the already high training cost of individual models, without explicitly controlling inter-alpha correlation. This raises a natural question: \textbf{\textit{can we simultaneously achieve efficient single-model prediction and diverse multi-alpha generation, without complex architectures or multi-stage pipelines?}}

To address this, we introduce \model{} (\textbf{M}ulti-\textbf{A}lpha \textbf{P}osition-aware \textbf{L}istwise \textbf{E}nsembling), a backbone-agnostic framework for efficient and diverse multi-alpha portfolio construction. 
\model{} replaces previous complex architectures with a unified prediction head that fuses an intra-stock path with a lightweight inter-stock attention path directly in prediction space. 
This head is further paired with an extreme-rank weighted listwise ranking loss that lets each alpha focus on the extreme-rank positions relevant to portfolio construction, and a diversity regularizer that explicitly penalizes pairwise correlation across alphas.
We further scale per-alpha capacity within both paths of the prediction head as the number of alphas grows, preventing the degradation observed in naive multi-alpha designs. 
Together, these components recover the classical alpha-mining diversity principle within a single model and a single training pass. \textbf{Our contributions are threefold:}

\begin{itemize}[leftmargin=*, itemsep=2pt, parsep=0pt]
    \item \textbf{A Unified, Capacity-Scaled Multi-Alpha Framework.}
    We propose \model{}, combining a unified prediction head, an extreme-rank weighted listwise ranking loss with a diversity regularizer, and a capacity-scaling scheme into a single model, without relying on separate models, specialized routing, or multi-stage pipelines to obtain diversity across multiple alphas.
    \item \textbf{State-of-the-Art Performance, Efficiency and Generalization.}
    Across four equity markets, \model{} achieves the best average Sharpe and Calmar ratios among nine baselines at a fraction of their parameter and training cost. Ablation studies confirm that each component contributes incremental gains, and \model{} generalizes consistently across five backbone architectures with fundamentally different inductive biases, with Sharpe and Calmar Ratio gains of 10--23\% and 17--43\%, respectively.
    \item \textbf{Mechanistic Evidence for Each Component.} Through complementary behavioral analyses, we show why each component works: the unified head already reduces inter-alpha correlation before any diversity loss is applied; anchoring the ranking loss within the extreme-rank region lets diversity regularization improve rather than erode per-alpha ranking quality; and capacity scaling prevents this balance from collapsing as more alphas are added, letting a single trained model match the diversification benefit of an explicit multi-model ensemble without its training cost.
\end{itemize}

%% file: content/prelim-new.tex




\section{Preliminary}
\label{sec:prelim}

\subsection{\textbf{Problem Formulation}}
Portfolio construction can be formulated as a learning-to-rank task across stocks using the future return as target. 
During training, given historical information $\bm{X} \in \mathbb{R}^{S \times T \times F}$ as input (i.e., $F$ features over $T$ past timesteps for each of the $S$ stocks), a model $f_{\theta}$ learns to predict the value or relative ranking of the $q$-day future return $\bm{y} \in \mathbb{R}^{S}$ for each stock, typically optimized with a pointwise, pairwise, or listwise objective: $\min_{\theta} \ \mathscr{L}\bigl(\bm{y},\, f_{\theta}(\bm{X})\bigr)$.
During inference, for each trading day $t$, a $q$-day long-only portfolio is formed by selecting the top-$k$ ranked stocks under $f_{\theta}$'s prediction, with portfolio weights $\bm{w}_t \in \mathbb{R}^{k}$ controlling the aggregation strength of each selected stock.
The portfolio performance is usually measured with Sharpe and Calmar Ratios, calculated from the daily return stream $\bm{p}_t = \bm{w}_t^{\intercal} \bm{C}_t$, where $\bm{C}_t \in \mathbb{R}^{k \times q}$ collects the realized returns of the selected stocks over the following $q$ days.

\subsection{Stock Prediction Architectures}
\label{sec:prelim-arch}
 
The model $f_\theta$ generally comprises three sequential components: a temporal encoder $t_\theta$ that extracts intra-stock representations from historical data, a cross-sectional module $g_\theta$ that captures relational context across stocks, and a prediction head $h_\theta$ that produces the final alpha signals.
 
\paragraph{\textbf{Temporal Encoder $\boldsymbol{t_\theta}$.}}
The temporal encoder processes each stock independently along the time axis, using architectures such as recurrent networks~\citep{chung2014empirical}, Transformers~\citep{vaswani2017attention}, or Mamba blocks~\citep{gu2023mamba}.
Given input $\bm{X} \in \mathbb{R}^{S \times T \times F}$, it produces an intra-stock hidden representation $\bm{H} = t_{\theta}(\bm{X}) \in \mathbb{R}^{S \times D}$, where $D$ is the hidden dimension. Since each stock is processed independently, cross-sectional structure is left to the downstream components.
 
\paragraph{\textbf{Cross-Sectional Module $\boldsymbol{g_\theta}$.}}
A stock's return is influenced not only by its own history but also by the behavior of related assets~\citep{fama1992cross}. The cross-sectional module addresses this by refining $\bm{H}$ through relational context aggregated across the stock universe, such that $\bm{H}^{\prime} = g_{\theta}(\bm{H}) \in \mathbb{R}^{S \times D}$. The function $g_{\theta}$ may be implemented using mechanisms such as inter-stock self-attention~\citep{yoo2021accurate,li2024master}, graph neural networks~\citep{hsu2021fingat}, and hypergraph-based methods~\citep{sawhney2021stock,xia2024cisthpan}. Regardless of the mechanism, the resulting representation $\bm{H}^{\prime}$ retains the same shape, $\mathbb{R}^{S \times D}$. Without cross-sectional modeling, $g_{\theta}$ reduces to the identity mapping, yielding $\bm{H}^{\prime}=\bm{H}$.

 
\paragraph{\textbf{Prediction Head $\boldsymbol{h_\theta}$.}}
The prediction head maps $\bm{H}^{\prime}$ to a scalar alpha score for each stock, yielding $\hat{\bm{y}} = h_{\theta}(\bm{H}^{\prime}) \in \mathbb{R}^{S}$. In most existing work, $h_\theta$ is a simple linear projection that produces a single ranking score per stock. \model{} redesigns both $g_\theta$ and $h_\theta$ as a unified module that produces diverse alpha signals within a single forward pass, as described in \cref{sec:alpha_attention_head}.

%% file: content/method.tex
\section{\model{}}

\begin{figure*}[ht]
    \centering
    \includegraphics[width=0.7\textwidth]{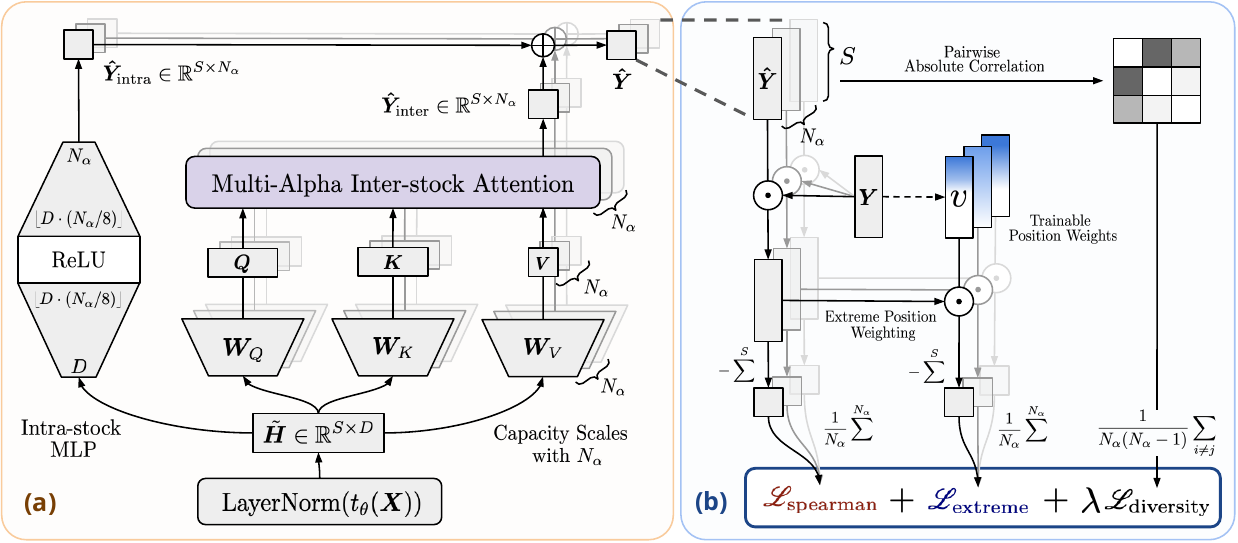}
    \caption{\textbf{\model{} overview.} \textbf{(a)} The unified head produces $N_\alpha$ signals. \textbf{(b)} Ranking and diversity losses combined as full objective.}
    \label{fig:model-struct}
\end{figure*}

\subsection{Multi-Alpha Generation Head}
\label{sec:alpha_attention_head}

Given the per-stock hidden representation $\bm{H} \in \mathbb{R}^{S \times D}$ produced by any backbone (\cref{sec:prelim-arch}), \model{} projects it to $N_{\alpha}$ alpha signals jointly in a single forward pass, an overview of \model{} architecture is provided in \cref{fig:model-struct}a. Unless otherwise specified, our experiments use a 2-layer Transformer with causal masking as the default temporal encoder $t_\theta$ (architectural details in Appendix~\ref{appendix:hyperparam}). We further verify generalization across additional backbones in \cref{sec:generalizability}.

The alpha generation head produces predictions via two paths that operate directly in prediction space rather than in hidden space, and whose outputs are summed to yield the final alpha signals.
Before entering the two paths, $\bm{H}$ is normalized via Layer Normalization~\citep{ba2016layer} over the hidden dimension, applied identically to every stock:
$\tilde{\bm{H}} = \mathrm{LayerNorm}(\bm{H}) \in \mathbb{R}^{S \times D}$, which serves as the shared input to both paths described next.

\paragraph{\textbf{Intra-Stock Path.}} The normalized representation $\tilde{\bm{H}}$ is projected to $N_{\alpha}$ alpha predictions via a two-layer MLP with ReLU activation:
\begin{equation}
    \bm{\hat{Y}}_{\text{intra}} = \mathrm{MLP}(\tilde{\bm{H}}) \in \mathbb{R}^{S \times N_{\alpha}},
\end{equation}
\noindent where the hidden dimension is set to $\lfloor D \cdot (N_\alpha / 8) \rfloor$, scaling proportionally with the number of alphas. This gives the model sufficient capacity to learn distinct intra-stock representations for each alpha, allowing different alphas to develop heterogeneous behaviors without interfering with each other's ranking quality.

\paragraph{\textbf{Inter-Stock Path.}} To inject relational context across the stock universe, we compute multi-head attention over the stock axis, assigning one head per alpha ($h = N_\alpha$) so that each alpha has its own independent set of query and key parameters. $\tilde{\bm{H}}$ is projected into query, key, and value spaces:
\begin{equation}
    \bm{Q} = \tilde{\bm{H}}\bm{W}_Q, \quad
    \bm{K} = \tilde{\bm{H}}\bm{W}_K, \quad
    \bm{V} = \tilde{\bm{H}}\bm{W}_V,
\end{equation}
\noindent where $\bm{W}_Q, \bm{W}_K \in \mathbb{R}^{D \times d_{\mathrm{emb}}}$, $\bm{W}_V \in \mathbb{R}^{D \times N_{\alpha}}$, and the embedding dimension is set to $d_{\mathrm{emb}} = \lfloor 2 \cdot D \cdot (N_\alpha / 8) \rfloor$. This scaling ensures that the per-head key dimension $d_k = d_{\mathrm{emb}} / N_\alpha$ remains approximately constant regardless of $N_\alpha$, so that adding more alphas does not reduce the capacity of each individual head. For each of the $N_\alpha$ heads, the $i$-th head output $\bm{A}^{(i)} \in \mathbb{R}^{S \times 1}$ is computed as:
\begin{equation}
    \bm{A}^{(i)} = \mathrm{softmax}\!\left(
        \frac{\bm{Q}^{(i)}{\bm{K}^{(i)}}^{\!\top}}{\sqrt{d_k}}
    \right)\bm{V}^{(i)},
\end{equation}
\noindent where $\bm{Q}^{(i)}, \bm{K}^{(i)} \in \mathbb{R}^{S \times d_k}$ and $\bm{V}^{(i)} \in \mathbb{R}^{S \times 1}$. Each head independently produces a scalar ranking score per stock, and the outputs are concatenated to form the inter-stock path output:
\begin{equation}
    \bm{\hat{Y}}_{\text{inter}} = \mathrm{Concat}\!\left(
        \bm{A}^{(1)}, \ldots, \bm{A}^{(N_\alpha)}\right) \in \mathbb{R}^{S \times N_{\alpha}}.
\end{equation}
\paragraph{\textbf{Prediction-level Residual Aggregation.}} Unlike a standard Transformer block, which applies the residual connection in hidden space followed by an FFN and a separate output head, the two paths here are summed in the output space to produce the final predictions:
\begin{equation}
    \bm{\hat{Y}} = \bm{\hat{Y}}_{\text{intra}} + \bm{\hat{Y}}_{\text{inter}}
    \;\in \mathbb{R}^{S \times N_{\alpha}}.
\end{equation}
This design empirically outperforms a full Transformer block across all backbones and markets (\cref{tab:component_ablation}). We further examine why this design is more effective in \cref{sec:attention_structure}, showing that it provides a favorable starting point for the diversity regularizer introduced later.
The resulting $\bm{\hat{Y}} \in \mathbb{R}^{S \times N_{\alpha}}$ contains $N_{\alpha}$ alpha signals. However, producing multiple outputs does not by itself guarantee diversity: without explicit encouragement, the $N_{\alpha}$ signals may converge to similar predictions, yielding a redundant ensemble that offers little diversification benefit~\citep{tulchinsky2019finding}. The following sections describe how we address this through a loss design that jointly optimizes each alpha's predictive quality (\cref{sec:method-loss-function}) and penalizes inter-alpha correlation (\cref{sec:method-alpha-ensemble}).

\subsection{Trainable Position-Aware Listwise Ranking}
\label{sec:method-loss-function}

Since portfolio construction depends on the relative ordering of assets rather than exact return values~\citep{10.1145/1273496.1273513,xia2008listwise,lan2014position,den2026integrating}, a natural choice is adopting a listwise objective that directly maximizes the Spearman rank correlation between predicted and target scores over all $S$ stocks:
\begin{equation}
\mathscr{L}_{\text{spearman}}
    = -\frac{1}{N_\alpha} \sum_{i=1}^{N_\alpha} \rho\!\left(\phi(\hat{\bm{y}}_i),\,\phi(\bm{y})\right),
\label{eq:spearman}
\end{equation}
\noindent where $\rho(\cdot,\cdot)$ is the Spearman rank correlation and $\phi(\cdot)$ 
is a differentiable rank surrogate~\citep{fang2019alpha}.
However, we find that optimizing a global Spearman objective conflicts with diversity regularization: because the diversity penalty must differentiate alphas across the entire ranking range, it disrupts rather than complements per-alpha ranking quality, leading to worse performance when the two are combined (see \cref{tab:component_ablation}).

\paragraph{\textbf{Extreme-Rank Weighted Spearman Loss.}}
To resolve this conflict, we concentrate each alpha's learning signal on the extreme-rank positions, so that diversity regularization only needs to differentiate alphas within this smaller, already-profitable region rather than across the full ranking range---an effect we verify in \cref{sec:loss_analysis}. 
We instantiate this as an extreme-rank profile that concentrates weight on stocks occupying extreme rank positions in the target $\bm{y}$, which is a natural choice since portfolio strategies typically act only on the extreme-rank positions~\citep{li2024master,sun2023mastering}.
Crucially, the weight of each stock is determined by its \emph{true rank} in $\bm{y}$ rather than a fixed position index, and is injected at the summation step of the Spearman correlation rather than before the rank surrogate computation. 

Specifically, each alpha $i \in \{1, \cdots ,N_\alpha\}$ has two learnable parameters for calculating the position weight: a sharpness $\xi^{(i)} > 0$ controlling the steepness of the transition between extreme and center positions, and a scale $\gamma^{(i)} \in (0,1)$ determining a margin $\delta^{(i)} = \gamma^{(i)} (S-1)/2$ that sets where the transition begins. 
As illustrated in \cref{fig:position_weight}, 
the resulting weight profile concentrates near 1 at the top-rank positions 
and near 0 at the center and below.
Allowing $\xi^{(i)}$ and $\gamma^{(i)}$ to be learned independently per alpha lets each alpha develop its own extreme-rank profile, which may further support diversity within this anchored region.
The full computation of the Extreme-Rank Weighted Spearman Loss is given in \cref{alg:extreme_rank}, yielding:
\begin{equation}
    \mathscr{L}_{\text{extreme}} = -\frac{1}{N_\alpha} \sum_{i=1}^{N_\alpha}
    C^{(i)} \sum_{s=1}^{S} \tilde{\phi}(\hat{\bm{y}}_i)_s \cdot \tilde{\phi}(\bm{y})_s \cdot v_s^{(i)},
\end{equation}
where $\tilde{\phi}(\cdot)$ denotes the demeaned and $\ell_2$-normalized rank surrogate vector, $v_s^{(i)}$ is the position weight of stock $s$ under alpha $i$, and $C^{(i)} = 1 + 2\delta^{(i)}/S$ is a coverage ratio that compensates for the reduced gradient magnitude when focusing on a narrow range of extreme positions, where we show the design concept in Appendix~\ref{appendix:ext_loss}.

Note that $\mathscr{L}_{\text{extreme}}$ concentrates the learning signal on extreme-rank positions but does not by itself enforce correct ordering elsewhere, we therefore retain $\mathscr{L}_{\text{spearman}}$ alongside it to preserve global ranking quality. The full training objective combining both terms with the diversity regularizer is given in \cref{sec:method-alpha-ensemble}.

\input{imgs/position_weighting_illustration}

\begin{algorithm}[t]
\caption{Extreme-Rank Weighted Spearman, PyTorch-style}
\label{alg:extreme_rank}
\begin{lstlisting}[style=pytorch, basicstyle=\ttfamily\footnotesize]
# y_hat: (N_alpha, S) predicted alpha scores
# y: (S,) target returns
# xi, gamma: (N_alpha,) learnable per-alpha params
# c: center index, (S-1)/2
 
def rank_surrogate(x): 
    return sigmoid(1.83 * (x - x.mean()) / (2 * x.std() + eps))
 
def normalize(phi): # demean and L2-normalize
    phi = phi - phi.mean()
    return phi / (phi.norm() + eps)
 
phi_y = normalize(rank_surrogate(y)) # (S,)
r = true_rank(y) # (S,) integer rank, r=0 for top stock
d = clip(c - r, min=0) # (S,) top-rank distance from center
 
L = 0
for i in range(N_alpha):
    phi_yi = normalize(rank_surrogate(y_hat[i])) # (S,)
    delta  = gamma[i] * c # transition margin for alpha i
    z = sigmoid(xi[i] * (d - delta)) # (S,) raw position activation
    v = z / (z.max() + eps) # (S,) position weight in [0,1]
    rho = (phi_yi * phi_y * v).sum() # position-weighted rank corr
    C = 1 + 2 * delta / S # coverage correction factor
    L = L + C * rho
loss = -L / N_alpha # negated mean loss over all alphas
\end{lstlisting}
\end{algorithm}

\subsection{Diversity-Controlled Multi-Alpha Ensemble}
\label{sec:method-alpha-ensemble}

To this end, we introduce a diversity regularizer during training and an
equal-weighted ensemble scheme across alphas at inference.
 
\paragraph{\textbf{Diversity Regularization.}}
We penalize redundancy among alpha heads by discouraging pairwise absolute correlation in their predicted signals:
\begin{equation}
    \mathscr{L}_{\text{diversity}} = \frac{1}{N_{\alpha}(N_{\alpha}-1)}
    \sum_{i \ne j} \left|\rho\!\left(\phi(\hat{\bm{y}}_i),\, \phi(\hat{\bm{y}}_j)\right)\right|,
\end{equation}
where $\rho(\cdot,\cdot)$ and $\phi(\cdot)$ are as in \cref{eq:spearman}. 
We use the absolute value because two strongly negatively correlated alphas select from opposite ends of the ranking, making one of them a poor alpha rather than a diversifier under a long-only constraint.

\paragraph{\textbf{Full Training Objective.}} As discussed in \cref{sec:method-loss-function}, $\mathscr{L}_{\text{spearman}}$ preserves global ranking quality while $\mathscr{L}_{\text{extreme}}$ emphasizes the extreme-rank positions most relevant to portfolio construction. The final training objective combines both per-alpha ranking losses with the diversity regularizer (\cref{fig:model-struct}b):
\begin{equation}
    \mathscr{L} = \mathscr{L}_{\text{spearman}} + \mathscr{L}_{\text{extreme}} + \lambda \mathscr{L}_{\text{diversity}},
\end{equation}
which we hereafter abbreviate as $\mathscr{L}_{\text{spr}}$, $\mathscr{L}_{\text{ext}}$, and $\mathscr{L}_{\text{div}}$, respectively.
Here, $\lambda > 0$ controls the trade-off between per-alpha predictive quality and inter-alpha diversity. Unless otherwise specified, we set $\lambda = 0.1$ throughout this paper.

\paragraph{\textbf{Alpha Aggregation.}}
At inference time, each alpha independently constructs its own portfolio by selecting its top-$k$ ranked stocks for each trading day (Appendix~\ref{appendix:eval_protocol}). The final portfolio return is the equal-weighted average across all $N_\alpha$ portfolios:
\begin{equation}
    p_t = \frac{1}{N_\alpha} \sum_{m=1}^{N_\alpha} p_t^{(m)},
\end{equation}
where $p_t^{(m)}$ denotes the return of the portfolio constructed by the $m$-th alpha at trading day $t$. This aggregation scheme is consistent with classical alpha mining practice, where each alpha operates independently and the ensemble benefit arises from signal diversity rather than explicit score fusion.

%% file: imgs/position_weighting_illustration.tex
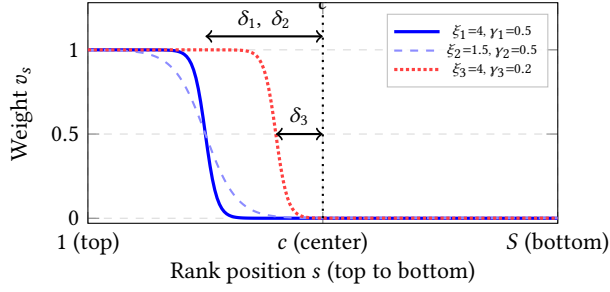
\begin{figure}[t]
\centering
\begin{tikzpicture}
\begin{axis}[
    width=0.92\columnwidth,
    height=4.5cm,
    xlabel={Rank position $s$ (top to bottom)},
    ylabel={Weight $v_s$},
    ylabel style={yshift=-10pt},
    xlabel style={yshift=4pt},
    xmin=0, xmax=20,
    ymin=-0.03, ymax=1.28,
    xtick={0, 10, 20},
    xticklabels={$1$ (top), $c$ (center), $S$ (bottom)},
    ytick={0, 0.5, 1},
    grid=major,
    grid style={dashed, gray!30},
    legend columns=1,
    legend style={
        at={(0.98, 0.95)},
        anchor=north east,
        font=\tiny,
        draw=gray!50,
        fill=white,
        fill opacity=0.85,
        text opacity=1,
        inner sep=2pt,
        row sep=-2pt,
    }
]
\addplot[blue, very thick, domain=0:20, samples=201]
    {1/(1+exp(-4*((10-x+abs(10-x))/2-5)))};
\addlegendentry{$\xi_1{=}4$, $\gamma_1{=}0.5$}
\addplot[blue!45, thick, dashed, domain=0:20, samples=201]
    {1/(1+exp(-1.5*((10-x+abs(10-x))/2-5)))};
\addlegendentry{$\xi_2{=}1.5$, $\gamma_2{=}0.5$}
\addplot[red!75, very thick, densely dotted, domain=0:20, samples=201]
    {1/(1+exp(-4*((10-x+abs(10-x))/2-2)))};
\addlegendentry{$\xi_3{=}4$, $\gamma_3{=}0.2$}
\draw[black, dotted, thick] (axis cs:10, 0) -- (axis cs:10, 1.22);
\node[black, font=\small, above] at (axis cs:10, 1.19) {$c$};
\draw[<->, thick, black] (axis cs:5, 1.08) -- (axis cs:10, 1.08);
\node[black, font=\small, above] at (axis cs:7.5, 1.08) {$\delta_1,\; \delta_2$};
\draw[<->, thick, black] (axis cs:8, 0.50) -- (axis cs:10, 0.50);
\node[black, font=\small, above] at (axis cs:9, 0.50) {$\delta_3$};
\end{axis}
\end{tikzpicture}
\caption{Position-aware weight $v_s$ shown across different sharpness $\xi$ and scale $\gamma$ values. Larger $\xi$ produces a sharper transition, while larger $\gamma$ shifts the margin $\delta$ further from the center, narrowing the range of top-ranked stocks that receive nonzero weight.}
\label{fig:position_weight}
\end{figure}

%% file: content/expr.tex
\input{tables/main_2}

\section{Experiments}
\label{sec:experiments}

\subsection{Experimental Setup}
\label{sec:experimental_setup}

\paragraph{\textbf{Datasets.}}
We evaluate \model{} across four major equity markets: CSI300 and CSI500 (China, 295 and 514 stocks), NI225 (Japan, 209 stocks), and SP500 (United States, 525 stocks). Following prior works~\citep{feng2019temporal,yoo2021accurate,sun2023mastering,den2026integrating}, we construct datasets using eight temporal features, including OHLCV and moving averages for each market. Models are trained on data from 2008--2019, validated on 2020, and tested on 2021--2024. Exact feature definitions and label formulations are provided in Appendix~\ref{appendix:features}.

\paragraph{\textbf{Baselines and Evaluation Metrics.}}
We compare \model{} against nine baselines spanning the architectural families discussed in \cref{sec:prelim-arch,sec:intro}: RankLSTM~\citep{feng2019temporal} (intra-stock only); FinGAT~\citep{hsu2021fingat}, MASTER~\citep{li2024master}, StockMixer~\citep{fan2024stockmixer}, and CI-STHPAN~\citep{xia2024cisthpan} (graph, attention, MLP-mixing, and hypergraph inter-stock mechanisms, respectively); MERA~\citep{liu2025mera} and DHMoE~\citep{chen2025dhmoe} (MoE routing, with DHMoE additionally using a diffusion module); AlphaMix~\citep{sun2023mastering} (explicit multi-model ensemble); and TIPS~\citep{den2026integrating} (multi-architecture distillation). We report Annual Return (AR), Sharpe Ratio (SR), and Calmar Ratio (CR), with metric definitions and evaluation protocol in Appendix~\ref{appendix:eval_protocol}. Hyperparameter and implementation details for baselines and \model{} are in Appendix~\ref{appendix:hyperparam} and~\ref{appendix:training_details}, respectively.


\subsection{Main Result}
\label{sec:perf-eval}
\cref{tab:main_results_2} reports the evaluation results across all four markets.
\model{} achieves the best average portfolio performance, with an average SR of 1.690 and CR of 2.175, outperforming every baseline on average---including the most recent MoE-based (MERA, DHMoE) and distillation-based (TIPS) methods---and attaining the highest SR in all four individual markets. On CR, \model{} leads on CSI300 and SP500 and is effectively tied with TIPS on NI225 (0.783 vs. 0.784), but trails both DHMoE and TIPS on CSI500.
Results with transaction costs are reported in Appendix~\ref{appendix:cost}.
This is achieved with only 186K parameters---an order of magnitude fewer than DHMoE (10.3M) and nearly 8x fewer than MERA (1.46M)---a training time of 3.60s per epoch that is roughly 2--2.5x faster than the three heaviest baselines (TIPS, MERA, CI-STHPAN), and inference-time FLOPs of just 0.130G, one to two orders of magnitude lower than attention- and MoE-based baselines (MASTER, MERA).
As discussed in \cref{sec:intro}, this confirms that architectural complexity does not translate into proportional performance gains, while the lighter baselines all trail \model{} by at least 0.23 SR---a gap that matters in practice, where portfolio models are frequently retrained on new market data.
The following sections examine which components of \model{} drive this performance (\cref{sec:component-ablation}) and whether these gains persist under scaling and across backbones (\cref{sec:scaling_law,sec:generalizability}). 

\input{tables/model_comp_ablation}

\subsection{Ablation Studies}
\label{sec:component-ablation}
\cref{tab:component_ablation} progressively builds from a vanilla intra-stock-only Transformer to the full \model{} design, organized into two groups, each broken down into labeled components (a)--(e) that we reference throughout this analysis.

\paragraph{\textbf{Group I: Lightweight intra-/inter-Stock Architecture}} Increasing number of alpha outputs (a) from 1 to 8 alone already improves performance, confirming that multi-alpha diversity has value even without further design changes. 
For inter-stock modeling (b), lightweight attention alone trades return for stability relative to the alternative (AR drops, while SR and CR rise); adding prediction-level aggregation on top of this lightweight attention partially recovers this trade-off and improves all three metrics, outperforming both the standard Transformer alternative and the intra-only design (a) in SR and CR, and confirming that inter-stock modeling adds value beyond intra-stock modeling alone.

\paragraph{\textbf{Group II: Diverse Alpha Ensemble through Single Model}}
For the loss design (c), combining vanilla Spearman correlation with the diversity regularizer degrades performance relative to Group I's final configuration, consistent with the conflict discussed in \cref{sec:method-loss-function}. Replacing it with our extreme-rank, position-aware loss instead improves all metrics, and adding diversity control further raises SR and CR despite a corresponding drop in AR (1.035 to 0.830). Scaling per-alpha capacity (d) in the attention and MLP paths yields further gains, consistent with a capacity bottleneck once alpha outputs must share the same backbone. Combining all components and scaling to 24 alphas (e) gives the full \model{} configuration, achieving the best overall performance.

\subsection{Generalizability Across Backbones}
\label{sec:generalizability}
\begin{figure}[t]
\centering
\includegraphics[width=\columnwidth]{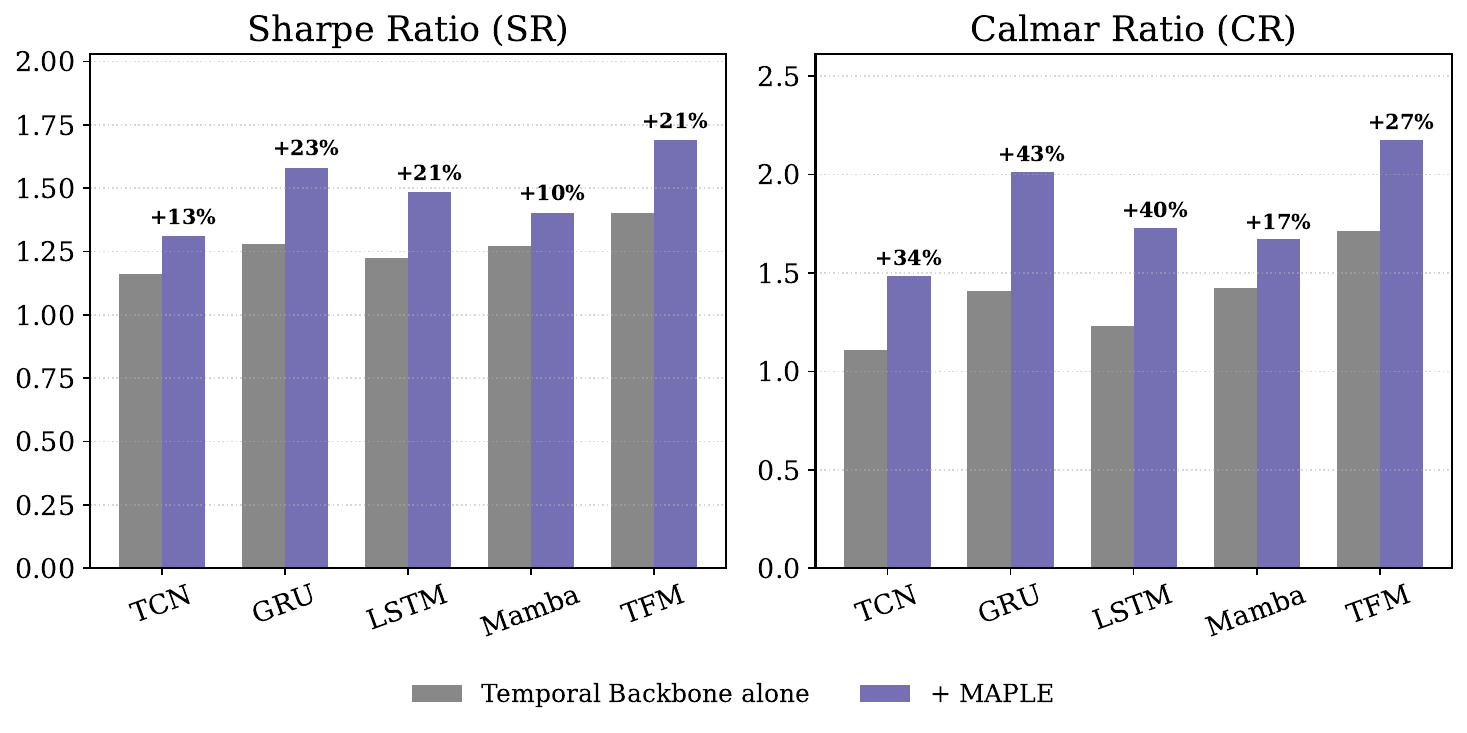}
\caption{Average Sharpe Ratio (left) and Calmar Ratio (right) across four
equity markets when replacing the intra-stock backbone in \model{}, compared
to each backbone alone (intra-stock-only, single alpha). Percentages denote
the relative improvement from adding \model{}.}
\label{fig:generalizability}
\end{figure}

\cref{fig:generalizability} evaluates whether \model{} generalizes beyond its default Transformer backbone by attaching it to four additional intra-stock encoders: TCN, GRU, LSTM, and Mamba. Across all five backbones, \model{} improves average SR and CR in every case, with SR gains ranging from 10\% (Mamba) to 23\% (GRU) and CR gains ranging from 17\% (Mamba) to 43\% (GRU). The best overall result is achieved when \model{} is paired with the Transformer backbone, but substantial gains are observed consistently across all five. 
These results indicate that \model{}'s benefits stem from principled design choices rather than architecture-specific tuning, and transfer reliably across backbones with fundamentally different inductive biases. Full per-market results are provided in Appendix~\ref{appendix:full_generalizability}.

%% file: tables/main_2.tex
\begin{table*}[t]
\caption{Main results across four equity markets. Portfolio performance is evaluated through SR and CR, while \textbf{Bold} and \underline{underline} denote the best and second best among all methods. For efficiency, we report both training-time and inference-time costs: model parameter count and training time per epoch for training; FLOPs, GPU memory usage, and inference time with a synthesized single-batch input matrix for inference. All efficiency metrics are lower-is-better.}
\centering
\resizebox{\textwidth}{!}{%
\begin{tabular}{lcc|cc|cc|cc|cc|cc|ccc}
\toprule
& \multicolumn{10}{c|}{\textbf{Portfolio Performance}} & \multicolumn{5}{c}{\textbf{Efficiency Performance}} \\
\textbf{Markets} & \multicolumn{2}{c}{\textbf{CSI300}} & \multicolumn{2}{c}{\textbf{CSI500}} & \multicolumn{2}{c}{\textbf{NI225}} & \multicolumn{2}{c}{\textbf{SP500}} & \multicolumn{2}{c|}{\textbf{Avg.}} & \multicolumn{2}{c}{\textbf{Training}} & \multicolumn{3}{c}{\textbf{Inference}} \\
\textbf{Metrics} $\rightarrow$ & SR & CR & SR & CR & SR & CR & SR & CR & SR & CR & Params & Tr. Time(s) & FLOPs(G) & Mem(MB) & Inf. Time(ms) \\
\midrule
RankLSTM & 0.753 & 0.663 & 0.869 & 0.699 & 0.610 & 0.447 & 1.414 & 2.138 & 0.912 & 0.987 & 52289 & 0.338 & 0.614 & 44.957 & 0.228 \\
FinGAT & 0.820 & 0.536 & 1.175 & 1.063 & 0.789 & 0.634 & 0.485 & 0.428 & 0.817 & 0.665 & 35686 & 2.140 & 0.025 & 44.249 & 3.370 \\
MASTER & 0.991 & 0.879 & 1.198 & 1.044 & 0.730 & 0.616 & 1.477 & 2.398 & 1.099 & 1.234 & 726601 & 1.479 & 8.725 & 72.210 & 1.343 \\
AlphaMix & 0.708 & 0.584 & 1.064 & 0.959 & 0.543 & 0.458 & 1.574 & 2.486 & 0.972 & 1.122 & 41257 & 0.715 & 0.482 & 36.268 & 1.182 \\
StockMixer & 0.726 & 0.612 & 0.615 & 0.421 & 0.450 & 0.361 & 0.919 & 0.899 & 0.678 & 0.573 & 14710 & 2.503 & 0.006 & 10.627 & 2.663 \\
CI-STHPAN & 0.180 & 0.126 & 0.425 & 0.290 & 0.767 & 0.607 & \underline{1.589} & 2.993 & 0.740 & 1.004 & 516868 & 9.001 & 2.475 & 22.096 & 22.541 \\
MERA & 1.075 & 1.055 & 0.677 & 0.472 & 0.769 & 0.554 & 1.117 & 1.131 & 0.910 & 0.828 & 1460017 & 8.932 & 11.153 & 61.719 & 7.909 \\
DHMoE & 0.961 & 0.977 & 1.716 & \textbf{2.490} & 0.679 & 0.585 & 0.980 & 0.680 & 1.084 & 1.222 & 10284722 & 4.499 & 1.571 & 55.642 & 6.455 \\
TIPS & \underline{1.343} & \underline{1.523} & \underline{2.010} & \underline{2.466} & \underline{0.958} & \textbf{0.784} & 1.506 & \underline{2.965} & \underline{1.454} & \underline{1.934} & 104897 & 7.236 & 0.810 & 30.670 & 0.760 \\
\midrule
\model{} & \textbf{1.851} & \textbf{2.062} & \textbf{2.161} & 1.961 & \textbf{0.991} & \underline{0.783} & \textbf{1.758} & \textbf{3.896} & \textbf{1.690} & \textbf{2.175} & 186480 & 3.601 & 0.130 & 38.142 & 1.685 \\ 
\bottomrule
\end{tabular}
}
\label{tab:main_results_2}
\end{table*}

%% file: tables/model_comp_ablation.tex
\begin{table}[t]
    \centering
    \caption{Component ablation (averaged across four markets) across two groups: a lightweight intra-/inter-stock architecture (Group I) and a diverse alpha ensemble built on Group I's final configuration (Group II). Within each group, \textit{Alternative} rows are compared against the preceding \textbf{Ours} row, while \textbf{Ours} rows accumulate progressively, each building on the one above. Bolded values improve on the preceding \textbf{Ours} row; \underline{underline} marks the best value in each column.}
    \resizebox{\columnwidth}{!}{%
    \begin{tabular}{lccc}
    \toprule
    \textbf{Component} & \textbf{AR} & \textbf{SR} & \textbf{CR} \\
    \midrule
    \multicolumn{4}{l}{\textbf{Group I --- Lightweight intra-/inter-Stock Architecture}} \\
    \midrule
    \emph{\textbf{(a) Intra-Stock Only: Single- vs. Multi-Alpha}} \\
    \textit{Baseline}: Vanilla Transformer ($N_\alpha=1$) & 0.865 & 1.402 & 1.713 \\
    \textbf{Ours}: + multi-alpha ($N_\alpha=8$) & \textbf{0.923} & \textbf{1.434} & \textbf{1.779} \\
    \midrule
    \emph{\textbf{(b) Inter-Stock Design}} \\
    \textit{Alternative}: Transformer Block & 1.038 & 1.305 & 1.511 \\
    \textbf{Ours}: + Lightweight Attn. (remove FFN, ${W}_V$ as pred.) & 0.509 & 1.414 & 1.676 \\
    \textbf{Ours}: + Prediction-level Residual Aggregation & \textbf{0.633} & \textbf{1.483} & \textbf{1.826} \\
    \midrule
    \midrule
    \multicolumn{4}{l}{\textbf{Group II --- Diverse Alpha Ensemble through Single Model}} \\
    \midrule
    \emph{\textbf{(c) Loss Design}} \\
    \textit{Alternative}: Vanilla Spearman + Diversity & 0.443 & 1.415 & 1.590 \\
    \textbf{Ours}: + Extreme-Rank Loss & \textbf{1.035} & \textbf{1.510} & \textbf{1.856} \\
    \textbf{Ours}: + Diversity Control & 0.830 & \textbf{1.579} & \textbf{1.996} \\
    \midrule
    \emph{\textbf{(d) Per-Alpha Capacity Scaling}} \\
    \textbf{Ours}: + Scaled Attention/MLP Capacity & \textbf{1.208} & \textbf{1.660} & \textbf{2.031} \\
    \midrule
    \emph{\textbf{(e) Alpha Count Scaling (Full Method)}} \\
    \textbf{Ours}: $N_\alpha = 24$ & \underline{\textbf{1.297}} & \underline{\textbf{1.690}} & \underline{\textbf{2.175}} \\
    \bottomrule
    \end{tabular}
    }
\label{tab:component_ablation}
\end{table}

%% file: content/analysis.tex
\section{Behavioral Analysis}

In this section, we examine why \model{}'s design choices are effective and how far their benefits extend. We show that \model{}'s gains do not come from architectural complexity or from aggressively suppressing correlation, but from three complementary mechanisms: an inter-stock structure that already provides a favorable starting point for diversity before any diversity loss is applied (\cref{sec:attention_structure}); a loss design that lets diversity regularization refine this starting point without eroding per-alpha ranking quality and portfolio return (\cref{sec:loss_analysis});
and a capacity-scaling scheme that sustains this balance not only as the number of alphas grows (\cref{sec:scaling_law}), but also across a wider range of the diversity weight $\lambda$ (\cref{sec:diversity_sensitivity}).
Unless stated otherwise, results in this section report averages across the four markets at $N_\alpha = 8$.

\input{tables/attention_analysis}

\input{tables/loss_analysis}

\subsection{Better Inter-stock Attention Structure}
\label{sec:attention_structure}

\cref{tab:attention_analysis} shows why our inter-stock attention structure already performs better before adding diversity-related design.
Given a model leveraging both intra- and inter-stock modules, $R^2_{\text{intra}}$ and $R^2_{\text{inter}}$ 
measure how much of the aggregated prediction is accounted for by each module's output alone, while Gradient Ratio compares the output-induced gradient norms of the two paths on the shared intra-stock backbone.

Without an inter-stock module, the prediction is determined entirely by the intra-stock path. 
Introducing a standard inter-stock module shifts gradient flow toward the inter-stock path, as expected (row 2, Gradient Ratio: 3.81).
Less expectedly, a prediction-level residual shifts it further still, leaving the prediction governed almost entirely by the inter-stock output (row 3, $R^2_\text{inter}$: 0.865 vs. $R^2_\text{intra}$: 0.271).
Our configuration (row 4) instead sets the $\bm{W}_V$ output dimension to directly match $N_\alpha$, 
letting it serve as the prediction under the same residual structure. This brings the gradient norms of the two paths much closer together.
Moreover, since $\bm{\hat{Y}} = \bm{\hat{Y}_\text{intra}} + \bm{\hat{Y}_\text{inter}}$,
the two coefficients sum to 1 only when the paths are uncorrelated; the sums are 1.008 for ours and 1.136 for the standard residual, indicating a shift from overlapping to nearly orthogonal contributions.

This shift in path dominance improves alpha diversity rather than individual alpha strength: relative to the intra-stock-only setting, our design trades a little per-alpha accuracy for a clear decrease in inter-alpha correlation (0.944 vs.\ 0.998), translating into a real diversification gain (0.071 vs.\ 0.003) and a higher ensemble SR (1.483 vs.\ 1.434), which is achieved even without explicitly optimizing for diversity. This gives a favorable starting point for the diversity-focused designs introduced next. 
Full gradient trajectories and the detailed calculations for this table are provided in Appendix~\ref{appendix:attention_analysis}.

\subsection{How Extreme-Rank enables Alpha Diversity}
\label{sec:loss_analysis}

\cref{tab:loss_analysis} shows why extreme loss ($\mathscr{L}_{\text{ext}}$) alone is insufficient, why naively combining spearman ($\mathscr{L}_{\text{spr}}$) with diversity regularization ($\mathscr{L}_{\text{div}}$) fails, and how combining all three resolves these issues. 

Used alone, $\mathscr{L}_{\text{ext}}$ pushes every alpha head toward the same top-ranked stocks (row 1): precision is the highest in the table, but the ensemble retains almost no diversification, yielding the most volatile returns and the lowest return-to-volatility ratio in the table.
$\mathscr{L}_{\text{ext}}$ is thus better suited as a regularizer on an existing ranking objective than as a standalone loss.
On the other hand, adding $\mathscr{L}_{\text{div}}$ to $\mathscr{L}_{\text{spr}}$ alone does reduce cross-alpha overlap and downside correlation (rows~2--3). 
However, it does so by spreading selections across a much wider set of names (union size 7.53 to 21.39) rather than concentrating on the subset that actually drives returns, with precision falling accordingly (0.223 to 0.203).
This is the failure mode discussed in \cref{sec:method-loss-function}, where diversity is gained at the expense of precision.

$\mathscr{L}_{\text{ext}}$ instead constrains where this diversification happens. Applying $\mathscr{L}_{\text{div}}$ on top of $\mathscr{L}_{\text{spr}} + \mathscr{L}_{\text{ext}}$ produces the same trends far more moderately, with precision dropping 5.1\% rather than 9.0\% (rows~4--5 vs.\ rows~2--3). This is because each alpha's learning signal already concentrated on the extreme-rank positions through $\mathscr{L}_{\text{ext}}$ before diversity is introduced, and $\mathscr{L}_{\text{div}}$ need only to differentiate alphas within this already-profitable region rather than across the full ranking range.
This lets diversity regularization improve rather than damage selection quality, yielding the highest return-to-volatility ratio for the full combination of all three losses.

\subsection{Scaling with Number of Alphas}
\label{sec:scaling_law}
\begin{figure}[t]
\centering
\includegraphics[width=\columnwidth]{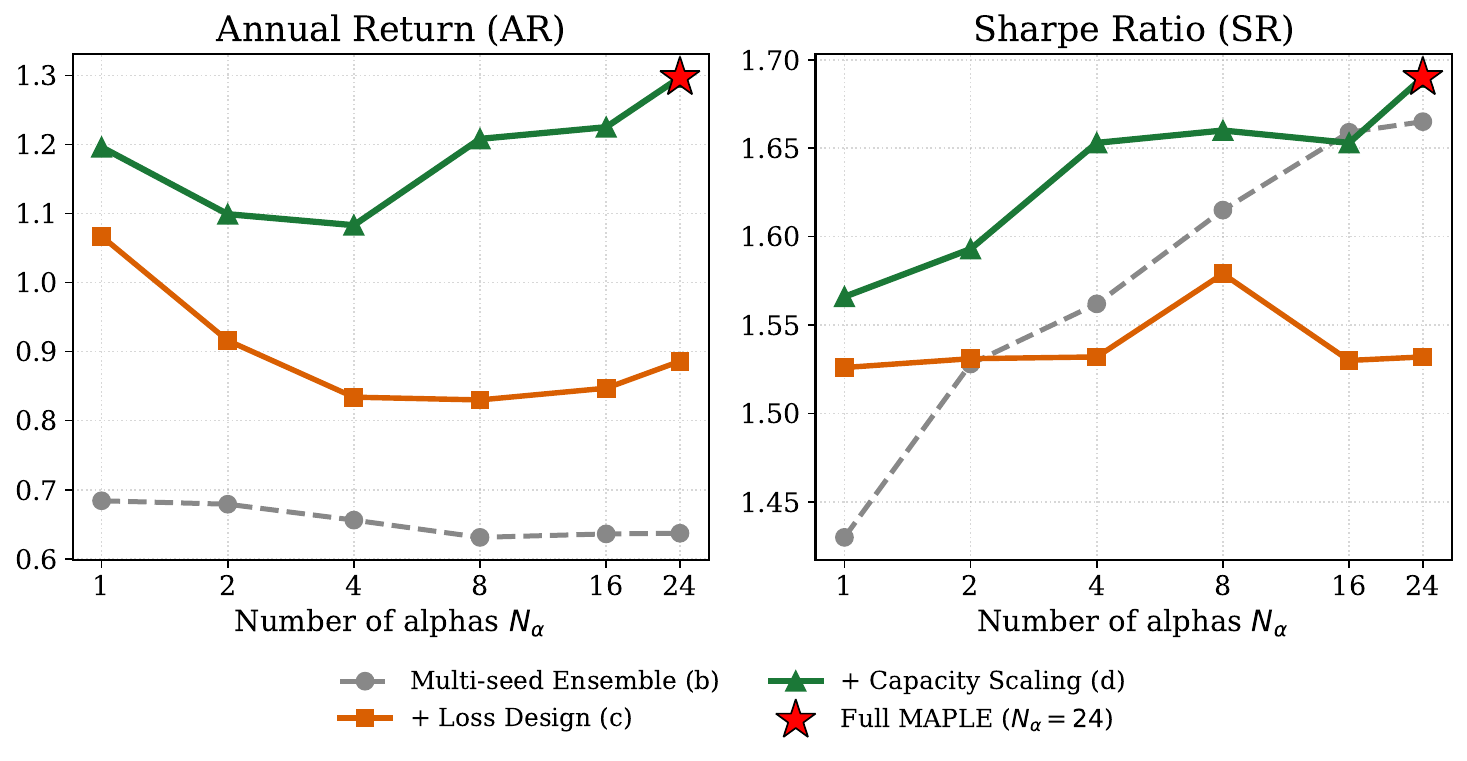}
\caption{Annual Return (left) and Sharpe Ratio (right) across different numbers of alphas $N_\alpha$, comparing loss design (c) and capacity-scaled configuration (d) against Multi-seed Ensemble under our Inter-Stock Design (b), as defined in \cref{tab:component_ablation}.}
\label{fig:alpha_scaling}
\end{figure}

\cref{fig:alpha_scaling} examines whether the balance between ranking quality and diversity established above holds as the number of alphas $N_\alpha$ grows, comparing our loss design and our capacity-scaled configuration on top of it (both defined in \cref{tab:component_ablation}).
As a reference point, we compare against Multi-seed Ensemble, which trains one instance of the inter-stock design per seed. Since its alphas never share capacity, it upper-bounds the diversity attainable at a given $N_\alpha$---at $N_\alpha \times$ the training and inference cost---and the question is how closely a single model can approach it.

Without capacity scaling, our loss design does not benefit from additional alphas. AR stays above the Multi-seed Ensemble reference but trends downward, while its SR stays within a narrow band and is overtaken at $N_\alpha$=4.
This reflects the capacity bottleneck discussed in \cref{sec:component-ablation}: shared capacity across alphas becomes insufficient as more diverse signals are demanded.
Adding capacity scaling addresses this bottleneck: AR recovers and increases beyond $N_\alpha=4$ while the other two configurations plateau or decline, and SR matches or surpasses Multi-seed Ensemble across most settings.
A \emph{single} trained model therefore matches an explicit ensemble's risk-adjusted performance while producing all $N_\alpha$ alphas in one forward pass, and at roughly twice the ensemble's annual return.
Full numerical results are provided in Appendix~\ref{appendix:full_scaling_results}.

These gains, however, depend on the alphas remaining diverse; We next examine how sensitive this balance is to the choice of $\lambda$, and whether capacity and alpha scaling make it more robust.

\subsection{Diminishing Returns with Diversity Weight}
\label{sec:diversity_sensitivity}

\begin{figure}[t]
\centering
\includegraphics[width=\columnwidth]{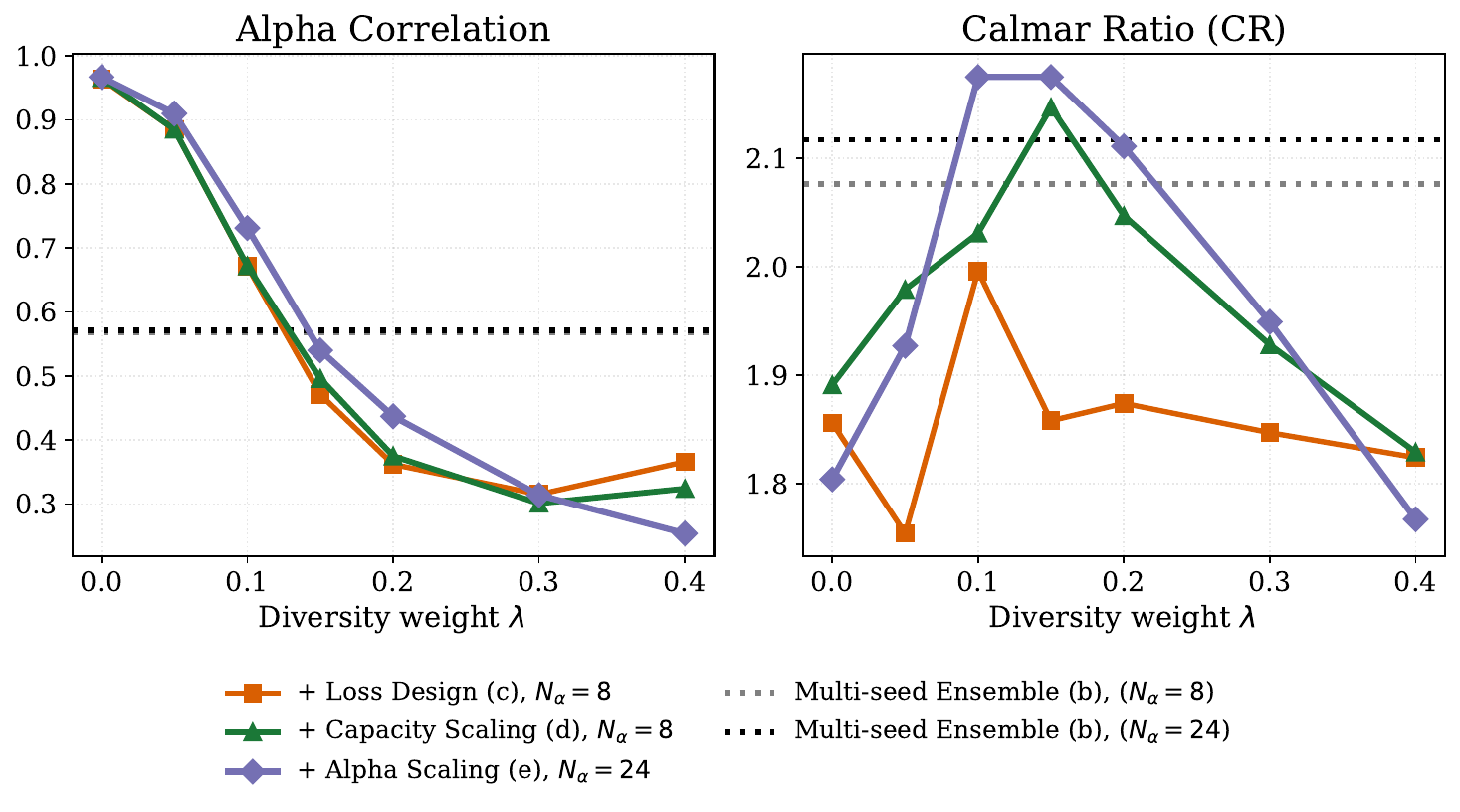}
\caption{Alpha Correlation (left) and Calmar Ratio (right) across different values of the diversity weight $\lambda$, comparing loss design (c, $N_\alpha=8$), capacity scaling (d, $N_\alpha=8$), and alpha scaling (e, $N_\alpha=24$), all defined in \cref{tab:component_ablation}. Dotted lines mark the corresponding values achieved by Multi-seed Ensemble under our Inter-Stock Design (b) at $N_\alpha=8$ and $N_\alpha=24$.}
\label{fig:diversity_sensitivity}
\end{figure}

\cref{fig:diversity_sensitivity} examines the effect of the diversity weight $\lambda$ on alpha correlation and risk-adjusted performance. At $\lambda = 0$, increasing $N_\alpha$ from 8 to 24 does not improve CR, confirming that more alpha outputs alone do not guarantee a diversification benefit. As $\lambda$ increases, alpha correlation decreases across all three configurations, confirming that the regularizer effectively reduces redundancy. CR itself first rises 
to a peak around $\lambda \in [0.1,0.15]$ and declines monotonically for $\lambda > 0.2$.
This indicates that beyond some point, the diversity term dominates training at the expense of per-alpha ranking quality, rather than continuing to improve returns.

This rise-then-decline shape reveals a further pattern. 
For our alpha-scaled configuration, CR plateaus while alpha correlation stays at or above the level Multi-seed Ensemble reaches---a level that emerges from independent training alone, with no diversity objective---and declines once a larger $\lambda$ pushes correlation below it. 
Independent training thus arrives at a correlation level that our regularizer should approach but not overshoot:
pushing beyond it only sacrifices per-alpha ranking quality without a corresponding gain.

Using the CR achieved by Multi-seed Ensemble as a reference point (gray and black dotted lines),
we find that capacity scaling and alpha scaling both bring CR to or above this reference 
at some $\lambda \in [0.1, 0.2]$ and stay within about ~2\% of it elsewhere in that range,
while the loss-only configuration never reaches it.
Scaling capacity and alpha count therefore make it substantially easier to identify and maintain a $\lambda$ that reaches this naturally diverse regime in practice. This lets a single model reproduce the diversification benefit that an explicit multi-model ensemble would otherwise require training separately to achieve. Full numerical results are provided in Appendix~\ref{appendix:full_sensitivity_results}.

%% file: tables/attention_analysis.tex
\begin{table*}[t]
    \centering
    \caption{Effect of the inter-stock prediction structure on gradient path dominance and alpha diversity. $R^2_{\text{intra}}$ and $R^2_{\text{inter}}$ measure how much of the aggregated prediction each module's output alone explains; Gradient Ratio is the inter-stock-to-intra gradient norm ratio. Correlation is the average pairwise alpha correlation and Individual SR the average per-alpha Sharpe ratio, with the diversification gain (Ensemble SR - Individual SR) in parentheses. Dashes mark configurations with no two-path decomposition.}
    \label{tab:attention_analysis}
    \resizebox{0.8\textwidth}{!}{%
    \begin{tabular}{ccccccccc}
    \toprule
    \multicolumn{3}{c}{\textbf{Inter-Stock Design}} & \multicolumn{3}{c}{\textbf{Path Dominance}} & \multicolumn{3}{c}{\textbf{Alpha Performance}} \\
    \cmidrule(lr){1-3} \cmidrule(lr){4-6} \cmidrule(lr){7-9}
    \textbf{Inter-stock} & \textbf{\(\bm{W}_V\) as Prediction} & \textbf{Pred-level Residual} & $\boldsymbol{R^2_{\text{intra}}}$ & $\boldsymbol{R^2_{\text{inter}}}$ & \textbf{Gradient Ratio} & \textbf{Correlation} ($\downarrow$) & \textbf{Individual SR} ($\uparrow$) & \textbf{Ensemble SR} ($\uparrow$)  \\
    \midrule
    \xmark & \xmark & \xmark & 1.000 & -- & -- & 0.998 & \textbf{1.431} & 1.434 (+0.003) \\
    \checkmarkb & \xmark & \xmark & -- & -- & 3.81 & 1.000 & 1.304 & 1.305 (+0.001) \\
    \checkmarkb & \xmark & \checkmarkb & 0.271 & 0.865 & 4.38 & 0.998 & 1.347 & 1.352 (+0.005) \\
    \checkmarkb & \checkmarkb & \checkmarkb & 0.586 & 0.422 & 1.20 & \textbf{0.944} & 1.412 & \textbf{1.483} (\textbf{+0.071}) \\
    \bottomrule
    \end{tabular}
    }
\end{table*}

%% file: tables/loss_analysis.tex
\begin{table*}[t]
    \centering
    \caption{Effects of extreme-rank and diversity losses on alpha behavior and portfolio performance. Overlap Ratio and Union Size describe pairwise/aggregate overlap among each alpha's top-5 stock selections; Downside Corr.\ is their return correlation on negative-return days. Precision and Recall are computed on the union of these selections against the 50 stocks with the highest realized 5-day returns. $\mu_r$, $\sigma_r$ are the mean and standard deviation of the 5-day cumulative returns of the selected baskets (in \%), while $\mu_r/\sigma_r$ is the return-to-volatility ratio.}
    \label{tab:loss_analysis}
    \resizebox{0.8\textwidth}{!}{%
    \begin{tabular}{ccccccccccc}
    \toprule
    \multicolumn{3}{c}{\textbf{Loss Components}} & \multicolumn{3}{c}{\textbf{Cross-alpha Dependency}} & \multicolumn{2}{c}{\textbf{Selection Quality}} & \multicolumn{3}{c}{\textbf{Performance}} \\
    \cmidrule(lr){1-3} \cmidrule(lr){4-6} \cmidrule(lr){7-8} \cmidrule(lr){9-11}
    \textbf{$\mathscr{L}_\text{spr}$} & \textbf{$\mathscr{L}_\text{ext}$} & \textbf{$\mathscr{L}_\text{div}$} & \textbf{Overlap Ratio} & \textbf{Union Size} & \textbf{Downside Corr.} & \textbf{Precision} & \textbf{Recall} & $\boldsymbol{\mu_r}$ ($\uparrow$) & $\boldsymbol{\sigma_r}$ ($\downarrow$) & $\boldsymbol{\mu_r/\sigma_r}$ ($\uparrow$) \\
    \midrule
    \xmark & \checkmarkb & \xmark & 0.930 & 5.920 & 0.988 & \textbf{0.299} & 0.035 & \textbf{0.49} & 5.8 & 0.084 \\
    \checkmarkb & \xmark & \xmark & 0.824 & 7.532 & 0.970 & 0.223 & 0.034 & 0.32 & \underline{3.2} & 0.100 \\
    \checkmarkb & \xmark & \checkmarkb & 0.317 & \textbf{21.386} & 0.809 & 0.203 & \textbf{0.085} & 0.22 & \textbf{2.2} & 0.100 \\
    \checkmarkb & \checkmarkb & \xmark & 0.902 & 6.312 & 0.980 & \underline{0.277} & 0.035 & \underline{0.47} & 4.5 & \underline{0.104} \\
    \checkmarkb & \checkmarkb & \checkmarkb & 0.680 & 10.368 & 0.920 & 0.263 & \underline{0.055} & 0.38 & 3.4 & \textbf{0.112} \\
    \bottomrule
    \end{tabular}
    }
\end{table*}

%% file: content/related.tex
\section{Related Works}

\paragraph{\textbf{Portfolio Construction with Deep Learning.}}
Deep learning methods for portfolio construction typically predict cross-sectional returns or ranking scores and form portfolios from the highest-ranked assets. Early studies primarily employed recurrent architectures to capture temporal market dynamics~\citep{feng2018enhancing,feng2019temporal}. Subsequent work extended this paradigm by explicitly modeling inter-stock dependencies through cross-stock attention~\citep{yoo2021accurate,li2024master,cao2024matcc}, relational graphs~\citep{hsu2021fingat,sawhney2021stock,xia2024cisthpan,zheng2023relational}, and MLP-based feature mixing~\citep{fan2024stockmixer,feng2025cryptomixer}. Beyond architectural advances, prior studies have explored pairwise ranking objectives~\citep{fang2019alpha}, behavior-informed stock representations~\citep{li2019individualized}, Transformer-based interaction modeling~\citep{wang2022adaptive}, volatility-aware learning~\citep{lin2025cspo}, and irrational-factor modeling~\citep{yang2025learning}. Despite their diversity, these approaches generally optimize a single predictive score for each stock, with performance improvements driven mainly by increasingly specialized model designs. In contrast, our framework extends portfolio construction from a single predictive signal to multiple complementary alphas learned within one model.

\paragraph{\textbf{Ensemble Learning in Stock Prediction.}}
Ensemble learning improves robustness by combining predictors that are both accurate and diverse~\citep{krogh1994neural,liu1999ensemble,hansen2002neural,lakshminarayanan2017simple}. This principle is particularly important in alpha mining, where combining low-correlated signals can substantially improve portfolio-level risk-adjusted returns~\citep{goldstein2015information,acharya2017measuring,tulchinsky2019finding}. Existing stock prediction ensembles typically obtain diversity by training multiple models with different random seeds or architectures~\citep{sun2023mastering,qin2026fineft}. Other approaches encourage specialization through routing or Mixture-of-Experts (MoE) mechanisms~\citep{yu2024miga,liu2025mera,chen2025dhmoe}, or consolidate multiple predictors through multi-stage distillation~\citep{den2026integrating}. However, these methods either require multiple training runs, introduce additional architectural complexity, or depend on separate training stages. More importantly, their diversity arises indirectly from initialization, model specialization, or training procedures, rather than from explicitly controlling the correlation among the resulting alpha signals. In contrast, our framework learns multiple complementary alphas within a single model and directly optimizes their inter-alpha diversity for portfolio construction.

%% file: content/conclude.tex
\section{Conclusion}

We presented \model{}, a framework for efficient and diverse multi-alpha generation built on three components: a unified prediction head that fuses an intra-stock path with a lightweight inter-stock attention path directly in prediction space, a loss design consisting of an extreme-rank weighted Spearman loss and a diversity regularizer, and a capacity-scaling scheme that sustains this design as the number of alphas grows.
Across four equity markets, \model{} achieves the best average Sharpe and Calmar ratios among nine baselines, including recent Mixture-of-Experts and distillation-based methods, 
at a fraction of their parameter and training cost;
it further generalizes consistently across five backbone architectures, confirming that these gains stem from \model{}'s own design rather than from any specific backbone. Our behavioral analysis explains why: the unified prediction head already reduces inter-alpha correlation before any diversity loss is applied, extreme-rank weighting lets the diversity regularizer improve rather than erode per-alpha ranking quality, and scaling per-alpha capacity sustains this balance as the number of alphas grows.
These results answer the question posed at the outset of this paper: efficient single-model prediction and diverse multi-alpha generation are not competing goals that require complex architectures or multi-stage training to reconcile. Our diversity regularizer currently decorrelates alphas conditioned on a single prediction target, asset class, and feature set, which bounds how much genuine diversity it can induce. Future work includes extending the model and loss design to automatically exploit diversity across these dimensions: treating different forecast horizons as distinct targets, extending beyond equities to asset classes such as futures, bonds, or cryptocurrencies, and incorporating richer feature sets such as financial statements or news beyond price and volume---for potentially larger diversity and more robust gains in risk-adjusted return.


%% file: content/appendix.tex
\section{Experimental Details}

\subsection{Definition of Input Features and Label}
\label{appendix:features}

For each stock and trading day, we construct an eight-dimensional temporal input vector consisting of five standardized OHLCV variables and three price-trend indicators. The open, high, low, close, and volume series are standardized separately using statistics estimated over the most recent 20 trading days. The close-price transformation is shown below as an example. To capture short- and medium-term price trends, we additionally use the relative deviation between the current closing price and its \(k\)-day moving average, where \(k\in{5,10,20}\). The prediction target is defined as the forward \(q\)-day return, with \(q=5\):

\begin{equation*}
\resizebox{\linewidth}{!}{$
z_{\text{close}}^t = \dfrac{x_{\text{close}}^t - \bar{x}_{\text{close}}^{[t-19:t]}}{\sigma_{x_{\text{close}}}^{[t-19:t]}}, \quad
z_{d_k}^t = \dfrac{\frac{1}{k}\sum_{i=0}^{k-1} x_{\text{close}}^{t-i}}{x_{\text{close}}^t} - 1, \quad
y_t = \dfrac{x_{\text{close}}^{t+q} - x_{\text{close}}^t}{x_{\text{close}}^t}
$}
\end{equation*}

\subsection{Hyperparameter Settings}
\label{appendix:hyperparam}

\paragraph{\textbf{Baselines.}}
For all baseline methods, we adopt the hyperparameter settings reported in the original papers or provided in the official implementations. Unless otherwise specified, we do not perform additional hyperparameter tuning.

\paragraph{\textbf{Intra-Stock Encoders.}}
\Cref{tab:temp_enc_hyperparam} shows the detailed hyperparameter settings of all intra-stock encoders we use in \cref{sec:generalizability}. For Transformer-based encoders, causal masking is applied along the temporal axis within each stock's independently processed sequence, preventing the model from attending to future time steps; stocks are processed independently by $t_\theta$, with cross-sectional interaction handled entirely by the inter-stock module $g_\theta$ (\cref{sec:prelim-arch}). For all backbones, the intra-stock representation $\bm{H}$ is taken as the hidden state at the final time step of this sequence.

\input{tables/tmp_enc_hyperparam}

\paragraph{\textbf{\model{}}}
\Cref{tab:maple_hyperparam} summarizes the hyperparameter settings of \model{}, organized by the temporal encoder $t_\theta$, the intra-stock MLP and inter-stock attention within the cross-sectional module $g_\theta$, and the loss function $\mathscr{L}$.
\input{tables/maple_hyperparam}

\subsection{Training \& Implementation Details}
\label{appendix:training_details}

All models are trained for 100 epochs with an effective batch size of 256, obtained by accumulating gradients over 4 steps with a per-step batch size of 64. For RankLSTM, \model{}, and intra-stock-only encoders in \cref{sec:generalizability}, we use the AdamW optimizer~\citep{loshchilov2017decoupled} with a learning rate of \(10^{-3}\), a weight decay of \(10^{-5}\), and gradient clipping with a maximum norm of 1.0. For the remaining baseline methods, we follow the optimizer and training configurations reported in their original papers or official implementations.

For each model, we conduct experiments using five random seeds: \(\{0,1,2,3,4\}\). For each seed, we select the checkpoint that achieves the best validation AR and report the mean performance across all seeds. All experiments are conducted on a single NVIDIA RTX 3090 GPU.

\subsection{Evaluation Protocol}
\label{appendix:eval_protocol}

We evaluate the predicted alpha scores using a staggered portfolio-construction protocol. Let
\(\bm{P} \in \mathbb{R}^{B \times S \times N_{\alpha}}\) denote the predicted scores over \(B\) trading days, \(S\) stocks, and \(N_{\alpha}\) alphas. Let \(\bm{R} \in \mathbb{R}^{B \times S}\)
denote the realized stock returns, where \(\bm{R}[d,s]\) is the realized return of stock \(s\) following formation day \(d\).

\paragraph{\textbf{Per-alpha sub-portfolios.}}
On each formation day \(d\), each alpha independently constructs a long-only sub-portfolio. For alpha \(a\), we select the \(k\) stocks with the highest predicted scores:
\begin{equation}
\mathcal{S}^{(d,a)} = \operatorname{arg\,top\text{-}k}_{s} \bm{P}[d,s,a],
\end{equation}
where \(k=5\). The selected stocks are assigned softmax-normalized weights:
\begin{equation}
w_{s}^{(d,a)} = \frac{\exp\!\left(\bm{P}[d,s,a]\right)}{\sum_{j \in \mathcal{S}^{(d,a)}}\exp\!\left(\bm{P}[d,j,a]\right)},\qquad s \in \mathcal{S}^{(d,a)}.
\end{equation}

The weights satisfy \(\sum_{s \in \mathcal{S}^{(d,a)}} w_{s}^{(d,a)} = 1\). Thus, stocks with larger predicted scores receive larger allocations within the corresponding sub-portfolio. For a portfolio formed on day \(d\), the return of alpha \(a\) at holding offset \(\tau\) is
\begin{equation}
r_{\tau}^{(d,a)} = \sum_{s \in \mathcal{S}^{(d,a)}} w_{s}^{(d,a)} \bm{R}[d+\tau,s], \qquad \tau = 0,\ldots,h_d-1,
\end{equation}
where \(h_d = \min(W,B-d)\) is the available holding length. Here, \(\tau=0\) denotes the first realized return following formation day \(d\). For a fixed formation day \(d\), the sub-portfolio returns are collected into \(\bm{r}^{\mathrm{sub}} \in \mathbb{R}^{N_{\alpha} \times h_d}\), where \(\bm{r}^{\mathrm{sub}}[a,\tau] = r_{\tau}^{(d,a)}\).

\paragraph{\textbf{Combination across alphas.}}
The alphas are combined at the portfolio level rather than at the score level. Each alpha first selects and weights its own top-\(k\) basket. Their resulting sub-portfolio returns are then averaged:

\begin{equation}
r_{\tau}^{(d)}
=
\frac{1}{N_{\alpha}}
\sum_{a=1}^{N_{\alpha}}
r_{\tau}^{(d,a)}.
\end{equation}

This construction preserves the stock-selection behavior of each alpha.
Different alphas may select different baskets, while a stock selected by
multiple alphas naturally receives greater aggregate exposure.

\paragraph{\textbf{Multiple rebalancing phases.}}
We set the holding and rebalancing interval to the prediction horizon \(W\).
To reduce sensitivity to the initial portfolio-formation date, we evaluate
\(W\) rebalancing phases. A portfolio formed on day \(d\) is assigned
to phase \(w = d \bmod W\). Portfolios in the same phase are therefore formed every \(W\) trading days. The holding-period returns generated from formation day \(d\) are written into the corresponding segment of that phase.

We store all phase-specific daily returns in a matrix \(\bm{Q} \in \mathbb{R}^{W \times B}\), the \(w\)-th row of \(\bm{Q}\) is denoted by \(\bm{q}^{(w)} = \bm{Q}[w,:]\), and represents the daily return sequence associated with rebalancing phase \(w\). Performance metrics are computed separately for each valid phase sequence \(\bm{q}^{(w)}\) and then averaged across all \(W\) phases.

\begin{algorithm}[H]
\caption{Portfolio Construction with Multiple Alphas}
\label{alg:portfolio_returns}
\begin{algorithmic}[1]
\Require Predicted scores
\(\bm{P} \in \mathbb{R}^{B \times S \times N_{\alpha}}\),
realized stock returns
\(\bm{R} \in \mathbb{R}^{B \times S}\),
basket size \(k\), and holding horizon \(W\)
\Ensure Phase-return matrix
\(\bm{Q} \in \mathbb{R}^{W \times B}\)

\State Initialize \(\bm{Q} \gets \bm{0}^{W \times B}\)

\For{\(d = 0\) to \(B-1\)}
    \State \(w \gets d \bmod W\)
    \Comment{Rebalancing phase}
    \State \(h \gets \min(W,B-d)\)
    \Comment{Available holding length}
    \State
    \(\bm{r}^{\mathrm{sub}}
    \gets
    \bm{0}^{N_{\alpha} \times h}\)

    \For{\(a = 0\) to \(N_{\alpha}-1\)}
        \State
        \(\bm{p} \gets \bm{P}[d,:,a]\)
        \Comment{Scores produced by alpha \(a\)}

        \State
        \(\mathcal{S}
        \gets
        \operatorname{arg\,top\text{-}k}(\bm{p})\)
        \Comment{Select the top-\(k\) stocks}

        \State
        \(\bm{w}
        \gets
        \operatorname{softmax}\!\left(\bm{p}[\mathcal{S}]\right)\)
        \Comment{Normalize the portfolio weights}

        \State
        \(\bm{r}^{\mathrm{sub}}[a,:]
        \gets
        \bm{R}[d:d+h,\mathcal{S}]\,\bm{w}\)
        \Comment{Holding-period returns of alpha \(a\)}
    \EndFor

    \State
    \(\bm{Q}[w,d:d+h]
    \gets
    \dfrac{1}{N_{\alpha}}
    \sum_{a=0}^{N_{\alpha}-1}
    \bm{r}^{\mathrm{sub}}[a,:]\)
    \Comment{Average the sub-portfolios}
\EndFor

\State \Return \(\bm{Q}\)
\end{algorithmic}
\end{algorithm}

\paragraph{\textbf{Performance metrics.}}
For phase \(w\), let

\begin{equation}
\bm{q}^{(w)}
=
\left[
q_{1}^{(w)},
q_{2}^{(w)},
\ldots,
q_{T_w}^{(w)}
\right]
\end{equation}

denote its valid daily portfolio-return sequence, where \(T_w\) is the number
of valid observations. Assuming \(252\) trading days per year, the annualized
return is \(\mathrm{AR}^{(w)}=252\,\bar{q}^{(w)}\), where \(\bar{q}^{(w)} = \frac{1}{T_w}\sum_{t=1}^{T_w} q_{t}^{(w)})\).

The annualized Sharpe ratio is
\begin{equation*}
\mathrm{SR}^{(w)}
=
\sqrt{252}\,
\frac{
    \bar{q}^{(w)}
}{
    \sigma\!\left(\bm{q}^{(w)}\right)
},
\end{equation*}
where \(\sigma\!\left(\bm{q}^{(w)}\right)\) denotes the standard deviation of
the daily portfolio returns. Following our implementation, the cumulative return sequence is computed by
summing the daily returns:
\begin{equation*}
V_{t}^{(w)} = \sum_{\tau=1}^{t} q_{\tau}^{(w)}.
\end{equation*}
The maximum drawdown is defined as the largest peak-to-trough decline in this
cumulative return sequence:
\begin{equation*}
\mathrm{MDD}^{(w)} = \max_{1 \leq t \leq T_w} \left(\max_{1 \leq \tau \leq t} V_{\tau}^{(w)} - V_{t}^{(w)}\right).
\end{equation*}
The Calmar ratio is
\begin{equation*}
\mathrm{CR}^{(w)} = \frac{\mathrm{AR}^{(w)}}{\mathrm{MDD}^{(w)}}.
\end{equation*}
Finally, each reported performance metric is averaged across the \(W\) rebalancing phases. For a generic metric \(M = \frac{1}{W} \sum_{w=0}^{W-1} M^{(w)}\). Higher values of AR, SR, and CR indicate better portfolio performance, whereas a lower MDD is preferred.

\section{Detail of Extreme Loss}
\label{appendix:ext_loss}

As the transition margin \(\delta_i\) increases, the position weights \(v_i\) become concentrated on a narrower subset of top-ranked stocks. Under a hard-threshold approximation of the sigmoid weighting, the effective number of contributing stocks decreases approximately linearly with \(\delta_i\). Consequently, the magnitude of the weighted correlation
\begin{equation*}
\rho_i=\sum_{s=1}^{S}\tilde{\phi}(\hat{\bm{y}}_{i})_s \cdot \tilde{\phi}(\bm{y})_s \cdot v^{(i)}_s,
\end{equation*}
also tends to decrease with the effective coverage. We therefore introduce
\begin{equation}
C_i = 1+\frac{2\delta_i}{S}
\label{eq:coverage}
\end{equation}
as a simple first-order coverage correction. Since \(S/2\) represents the approximate scale of the ranked region, the normalized margin \(2\delta_i/S\) measures how far the weighting has shifted from its baseline coverage. Thus, \(C_i\) increases linearly from \Cref{eq:coverage} as the selected region becomes narrower. This correction is heuristic rather than an exact normalization, and is intended to reduce differences in loss scale caused solely by different learned coverage widths.

\section{Detailed Experiments}
\subsection{Performance Degradation with Transaction Costs}
\label{appendix:cost}
We model transaction costs covering brokerage commissions, exchange fees, and regulatory charges: CSI300 \& CSI500 incurs $\sim$0.006\% (buy) and $\sim$0.056\% (sell), NI225 $\sim$0.002\% on both sides, and the SP500 negligible costs ($\sim$0\%). With 5-day rebalancing, annualized costs amount to 3.12\%, 0.20\%, and 0\% respectively, is the upper bound of transaction cost degrade based on 100\% turnover rate. \Cref{tab:main_results_cost} shows the performance considering transaction costs. Compare with \Cref{tab:main_results_2}, \model{} maintains the best average performance with least transaction cost degradation.
\input{tables/cost}

\subsection{Detail results of \cref{sec:generalizability}}
\label{appendix:full_generalizability}

\input{tables/generalizability}
\Cref{tab:generalizability} reports the results across four equity markets and five intra-stock backbones. 
Overall, \model{} improves both metrics on 15 of the 20 backbone–market pairs and at least one metric on 19 of 20, with TCN on CSI300 the sole exception; the four mixed cases trade one metric against the other rather than degrading both.
Nevertheless, when averaged across markets, \model{} improves both SR and CR for every backbone, demonstrating its broad compatibility with different temporal architectures and its consistent gains in performance.

\section{Detailed Analysis Calculations \& Results}

\subsection{Path-dominance \texorpdfstring{$R^2$}{R2} \& Gradient Competition Trajectory}
\label{appendix:attention_analysis}

\paragraph{\textbf{Path-dominance \texorpdfstring{$R^2$}{R2}}}
We use \texttt{sklearn.metrics.r2\_score}~\citep{pedregosa2011scikit} to calculate \(R^2_{\mathrm{intra}}\) and \(R^2_{\mathrm{inter}}\), which denotes the \(R^2\) scores between \((\hat{\bm{Y}}_{\mathrm{intra}}, \hat{\bm{Y}}\)) and \((\hat{\bm{Y}}_{\mathrm{inter}}, \hat{\bm{Y}}\)), respectively.

\paragraph{\textbf{Gradient Competition Trajectory}}

\begin{figure}[H]
    \centering
    \includegraphics[width=0.9\linewidth]{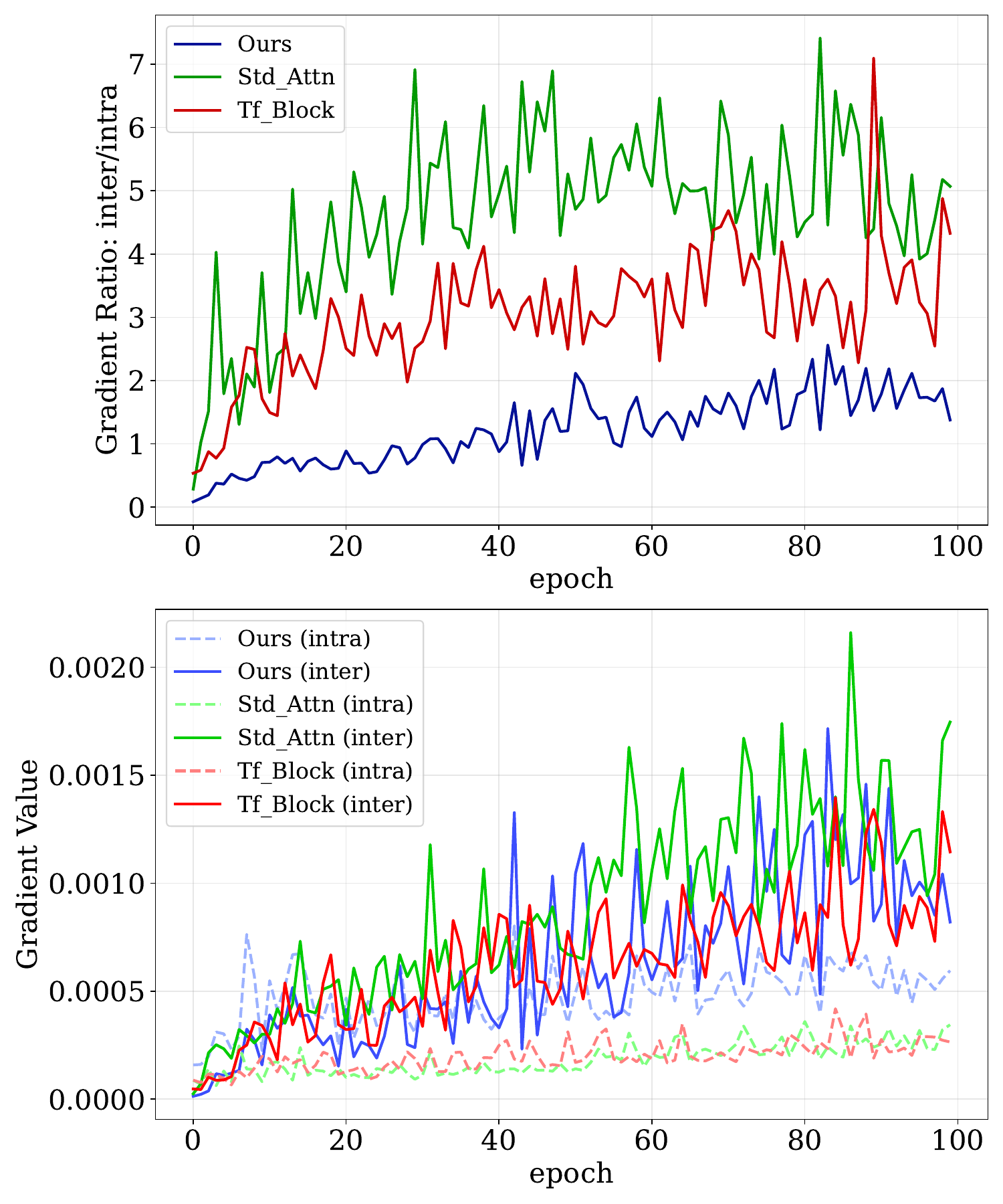}
    \caption{Gradient trajectory of different attention structure through 100 epochs. Top: competition ratio of inter-stock gradient and intra-stock gradient trajectories. Bottom: Separate inter-stock and intra-stock gradient trajectories.}
    \label{fig:gradient_trajectory}
\end{figure}

\Cref{fig:gradient_trajectory} shows the full gradient trajectory of intra-stock output \(\bm{\hat{Y}}_{\text{intra}}\) and inter-stock output \(\bm{\hat{Y}}_{\text{inter}}\). We probe the model once per epoch measuring the RMS gradient each path induces on the shared backbone \(t_{\theta}\) in isolation. We report the competition ratio:
\begin{equation*}
\rho = \dfrac{|\nabla \hat{\bm{Y}}_{\text{inter}}/\nabla t_{\theta}|}{|\nabla \hat{\bm{Y}}_{\text{intra}}/\nabla t_{\theta}|},
\end{equation*}
where \(\rho > 1\) indicates the inter-stock path exerts more pull on shared representation capacity than the intra-stock path.

All three architectures dominate by intra-stock at the very beginning training epochs, and shift toward inter-stock dominance as training proceeds, but they separate sharply in both the timing and the magnitude of that shift. As a result, Ours architecture reaches an equilibrium in which the two paths pull on the backbone at comparable strength, while \texttt{Tf\_Block} and \texttt{Std\_Attn} converges to a regime where the cross-stock path dominates the shared representation by 4-5 times higher. Two variants: \texttt{Tf\_Block} and \texttt{Std\_Attn} are the second and third row in \Cref{tab:attention_analysis}, respectively. 

\subsection{Loss Analysis}
\label{appendix:loss_analysis}

\begin{figure}[H]
    \centering
    \includegraphics[width=\linewidth]{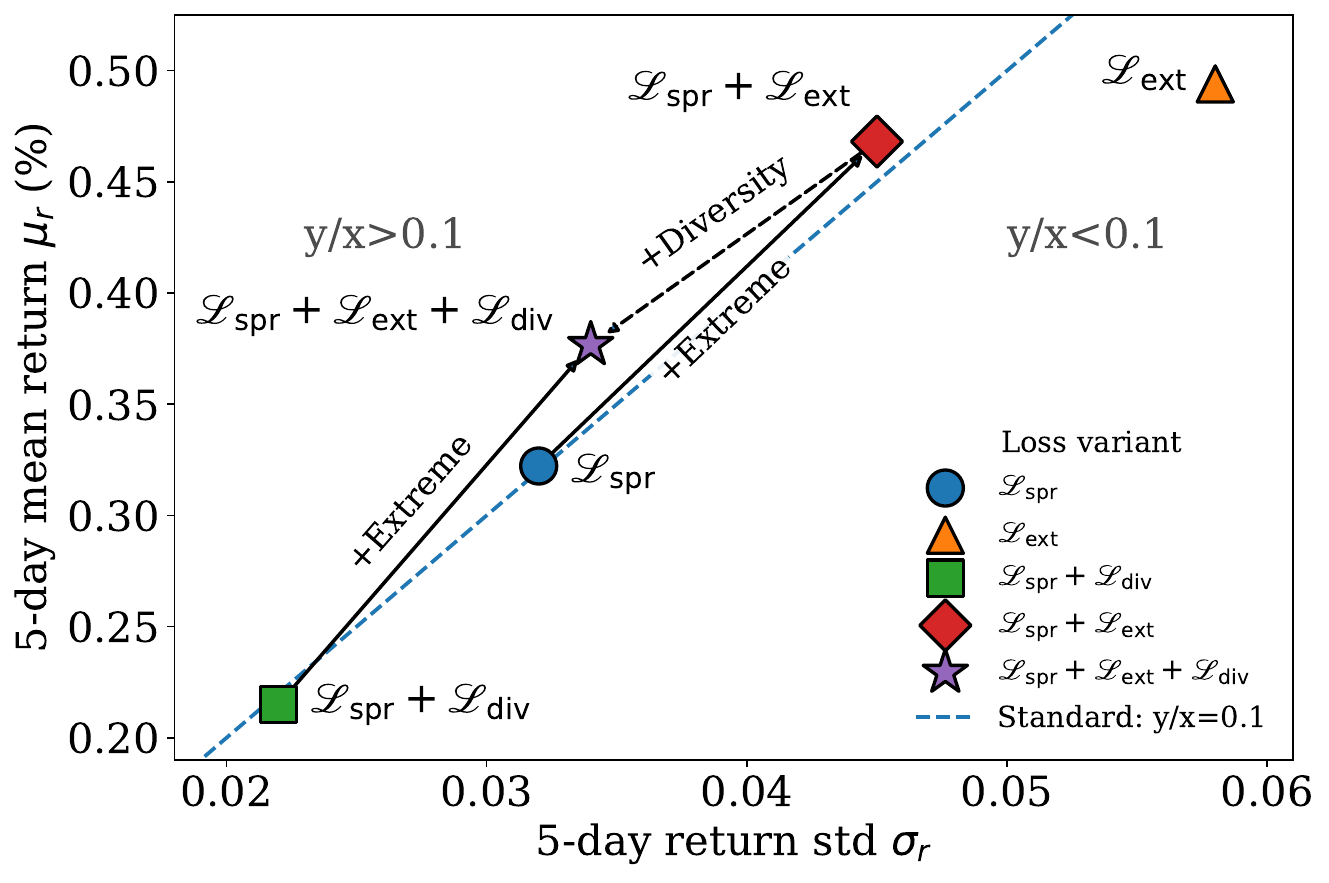}
    \caption{Ablation of the extreme loss $\mathscr{L}_{\text{ext}}$ and the diversity regularizer $\mathscr{L}_{\text{div}}$. Each term isolates one side of the risk--return trade-off: $\mathscr{L}_{\text{ext}}$ improves mean return but inflates volatility, while $\mathscr{L}_{\text{div}}$ reverses both. Their combination attains the highest risk-adjusted score, indicating the two objectives are complementary rather than redundant.}
    \label{fig:loss_tradeoff}
\end{figure}

\Cref{fig:loss_tradeoff} plots each loss variant in the risk–return plane spanned by the 5-day return standard deviation \(\sigma_r\) (horizontal) and mean return \(\mu_r\) (vertical, in \%), where the ratio \(\mu_r/\sigma_r\) acts as a risk-adjusted score and the dashed line \(\mu_r/\sigma_r = 0.1\) marks a fixed-efficiency reference: points to its upper-left earn more return per unit of risk, points to its lower-right earn less. The two auxiliaries move the strategy in opposite and complementary directions. The extreme loss \(\mathscr{L}_{\text{ext}}\) is a return amplifier that trades away efficiency: adding it shifts the strategy up-and-right along both axes: \(\mathscr{L}_{\text{spr}} \to \mathscr{L}_{\text{spr}}+\mathscr{L}_{\text{ext}}\) raises mean return by $+0.146\%$ but also inflates volatility by $+0.013$, and the same displacement appears in the presence of diversity ($\mathscr{L}_{\text{spr}}+\mathscr{L}_{\text{div}} \to \mathscr{L}_{\text{spr}}+\mathscr{L}_{\text{ext}}+\mathscr{L}_{\text{div}}$, $+0.162\%$ return, $+0.012$ risk), consistent with its role of sharpening the tails of the score distribution to concentrate exposure on extreme-ranked names. Left unchecked this pushes the strategy toward the high-risk regime occupied by the extreme-only model $\mathscr{L}_{\text{ext}}$ ($\sigma_r=0.058$), which posts the largest raw return ($\mu_r=0.494\%$) but the worst efficiency of all five variants ($\mu_r/\sigma_r=0.085$, well below the reference line). The diversity loss is the countervailing risk regulator: the $\mathscr{L}_{\text{spr}}+\mathscr{L}_{\text{ext}} \to \mathscr{L}_{\text{spr}}+\mathscr{L}_{\text{ext}}+\mathscr{L}_{\text{div}}$ arrow points down-and-left, shedding $-0.011$ of volatility while giving back only $-0.091\%$ of return — cutting risk proportionally harder than return and thereby lifting the point back across the efficiency frontier. Acting together, the two auxiliaries relocate the deployed model $\mathscr{L}_{\text{spr}}+\mathscr{L}_{\text{ext}}+\mathscr{L}_{\text{div}}$ to the best risk-adjusted position in the plane ($\mu_r/\sigma_r = 0.111$, the highest of any variant and above the reference line, versus $0.101$ for $\mathscr{L}_{\text{spr}}$ and $0.085$ for $\mathscr{L}_{\text{ext}}$): the extreme loss supplies the return, and the diversity loss — realized through the decorrelated alpha heads — converts the resulting volatility back into a controlled-risk profile. This decomposition is the mechanistic counterpart to the headline result that neither auxiliary alone is sufficient; the risk-control benefit that matters operationally (lower volatility and downstream MDD) is specifically what the diversity term buys back from the return-seeking extreme term.

\subsection{Full Numerical Results of \cref{sec:scaling_law}}

\label{appendix:full_scaling_results}

\input{tables/alpha_scaling}

\subsection{Full Numerical Results of \cref{sec:diversity_sensitivity}}
\label{appendix:full_sensitivity_results}
\input{tables/diversity_sensitivity}

%% file: tables/tmp_enc_hyperparam.tex
\begin{table}[H]
    \centering
    \caption{Intra-stock encoder hyperparameter configurations.}
    \resizebox{0.45\textwidth}{!}{%
    \begin{tabular}{lccccc}
    \toprule
    \textbf{Model} & \textbf{Hidden} & \textbf{Layers} & \textbf{FFN} & \textbf{Heads} & \textbf{Dropout} \\
    \midrule
    GRU / LSTM          & 64 & 2 & —   & — & 0.0 \\
    TCN$^*$        & 64 & 2 & —   & — & 0.0 \\
    Transformer$^\dagger$        & 64 & 2 & 256 & 8 & 0.1 \\
    Mamba$^\ddagger$          & 32 & 1 & 64  & — & 0.1 \\
    \bottomrule
    \end{tabular}}
    \begin{minipage}{0.45\textwidth}
    \footnotesize
    $^*$ TCN: kernel size 4. \\
    $^\dagger$ Transformer: FFN activation is ReLU. \\
    $^\ddagger$ Mamba: state size 16, conv size 4, expand factor 2\\
    \end{minipage}
    \label{tab:temp_enc_hyperparam}
\end{table}

%% file: tables/maple_hyperparam.tex

\begin{table}[h]
    \centering
    \caption{\model{} hyperparameter configurations}
    \begin{tabular}{lc}
    \toprule
    \textbf{Configuration} & \textbf{Value} \\
    \midrule
    \multicolumn{2}{l}{\emph{\textbf{General Settings}}} \\
    Intra-stock encoder $t_\theta$ & Transformer \\
    Number of Alphas $N_{\alpha}$ & 24 \\
    \midrule
    \multicolumn{2}{l}{\emph{\textbf{Intra-Stock MLP}}} \\
    hidden size & \(\lfloor D \cdot (N_{\alpha}/8)\rfloor\) \\
    \midrule
    \multicolumn{2}{l}{\emph{\textbf{Inter-Stock Attention}}} \\
    \(\bm{W}_Q, \bm{W}_K\) hidden & \(\lfloor2 \cdot D\cdot(N_{\alpha}/8)\rfloor\) \\
    \(\bm{W}_V\) hidden & \(N_{\alpha}\) \\
    Attention Heads & \(N_{\alpha}\) \\
    Dropout & 0.1 \\
    \midrule
    \multicolumn{2}{l}{\emph{\textbf{Loss Function}}} \\
    Init. value for Trainable Sharpness (\(\xi\)) & 10 \\
    Init. value for Trainable Margin (\(\gamma\)) & 0.8 \\
    Diversity weight (\(\lambda\)) & 0.1 \\
    \bottomrule
    \end{tabular}
    \label{tab:maple_hyperparam}
\end{table}

%% file: tables/cost.tex
\begin{table}[h]
\caption{Main result considering transaction costs}
\centering
\resizebox{\linewidth}{!}{%
\begin{tabular}{lcc|cc|cc|cc|cc}
\toprule
\textbf{Markets} & \multicolumn{2}{c}{\textbf{CSI300}} & \multicolumn{2}{c}{\textbf{CSI500}} & \multicolumn{2}{c}{\textbf{NI225}} & \multicolumn{2}{c}{\textbf{SP500}} & \multicolumn{2}{c}{\textbf{Avg.}} \\
\textbf{Metrics} $\rightarrow$ & SR & CR & SR & CR & SR & CR & SR & CR & SR & CR \\
\midrule
RankLSTM & 0.658 & 0.548 & 0.792 & 0.605 & 0.603 & 0.441 & 1.414 & 2.138 & 0.867 & 0.933 \\
FinGAT & 0.713 & 0.466 & 1.070 & 0.968 & 0.781 & 0.627 & 0.485 & 0.428 & 0.762 & 0.622 \\
MASTER & 0.811 & 0.719 & 1.019 & 0.888 & 0.721 & 0.608 & 1.477 & 2.398 & 1.007 & 1.153 \\
AlphaMix & 0.588 & 0.485 & 0.960 & 0.866 & 0.536 & 0.452 & 1.574 & 2.486 & 0.915 & 1.072 \\
StockMixer & 0.429 & 0.362 & 0.467 & 0.320 & 0.440 & 0.353 & 0.919 & 0.899 & 0.564 & 0.484 \\
CI-STHPAN & 0.036 & 0.025 & 0.352 & 0.240 & 0.763 & 0.604 & \underline{1.589} & 2.993 & 0.685 & 0.966 \\
MERA & 0.967 & 0.949 & 0.614 & 0.428 & 0.761 & 0.548 & 1.117 & 1.131 & 0.865 & 0.764 \\
DHMoE & 0.709 & 0.721 & 1.458 & \underline{2.115} & 0.669 & 0.577 & 0.980 & 0.680 & 0.954 & 1.023 \\
TIPS & \underline{1.260} & \underline{1.428} & \underline{1.949} & \textbf{2.392} & \underline{0.952} & \textbf{0.779} & 1.506 & \underline{2.965} & \underline{1.417} & \underline{1.891} \\
\midrule
\model{} & \textbf{1.786} & \textbf{1.990} & \textbf{2.118} & 1.922 & \textbf{0.985} & \underline{0.778} & \textbf{1.758} & \textbf{3.896} & \textbf{1.662} & \textbf{2.146} \\
\bottomrule
\end{tabular}
}
\label{tab:main_results_cost}
\end{table}

%% file: tables/generalizability.tex
\begin{table}[H]
    \caption{Generalizability on replacing different intra-stock backbones in \model{} across four equity markets. \textbf{Bold} indicate that \model{} performs better than corresponding backbone baseline (intra-stock-only, single alpha). \underline{Underline} denotes the best among all methods.}
    \centering
    \resizebox{\columnwidth}{!}{%
        \begin{tabular}{lcc|cc|cc|cc|cc}
        \toprule
        \textbf{Markets} & \multicolumn{2}{c}{\textbf{CSI300}} & \multicolumn{2}{c}{\textbf{CSI500}} & \multicolumn{2}{c}{\textbf{NI225}} & \multicolumn{2}{c}{\textbf{SP500}} & \multicolumn{2}{c}{\textbf{Avg.}} \\
        \textbf{Metrics} $\rightarrow$ & SR & CR & SR & CR & SR & CR & SR & CR & SR & CR \\
        \midrule
        TCN & 1.391 & 1.304 & 1.612 & 1.474 & 0.559 & 0.434 & 1.076 & 1.219 & 1.160 & 1.108 \\
        \textbf{+\model{}} & 
        1.235 & 1.129 & \textbf{1.789} & \textbf{1.764} & \textbf{0.838} & \textbf{0.679} & \textbf{1.382} & \textbf{2.366} & \textbf{1.311} & \textbf{1.484} \\
        \midrule
        GRU & 1.126 & 1.051 & 1.611 & 1.474 & 0.787 & 0.581 & 1.594 & 2.516 & 1.280 & 1.406 \\
        \textbf{+\model{}} & 
        \textbf{1.641} & \textbf{1.696} & \textbf{1.937} & \textbf{1.860} & \textbf{1.088} & \textbf{0.805} & \textbf{1.654} & \textbf{3.692} & \textbf{1.580} & \textbf{2.013} \\
        \midrule
        LSTM & 1.139 & 0.934 & 1.362 & 1.117 & 0.789 & 0.602 & 1.608 & 2.276 & 1.225 & 1.232 \\
        \textbf{+\model{}} & 
        \textbf{1.471} & 
        \textbf{1.441} &  
        \textbf{1.740} & 
        \textbf{1.459} & 
        \underline{\textbf{1.191}} & 
        \underline{\textbf{0.857}} & 
        1.539 & 
        \textbf{3.151} & 
        \textbf{1.485} & 
        \textbf{1.727} \\
        \midrule
        Mamba & 1.071 & 1.017 & 1.389 & 1.251 & 0.940 & 0.762 & 1.687 & 2.669 & 1.272 & 1.425 \\
        \textbf{+\model{}} & 
        \textbf{1.193} & 
        \textbf{1.197} & 
        \textbf{1.753} & 
        \textbf{1.577} & 
        \textbf{0.996} & 
        0.701 &  
        1.671 &
        \textbf{3.210} & 
        \textbf{1.403} & 
        \textbf{1.671} \\
        \midrule
        TFM & 1.509 & 1.411 & 2.006 & \underline{2.122} & 0.648 & 0.510 & 1.447 & 2.811 & 1.402 & 1.713 \\
        \textbf{+\model{}} & 
        \underline{\textbf{1.851}} & 
        \underline{\textbf{2.062}} &  
        \underline{\textbf{2.161}} & 
        1.961 & 
        \textbf{0.991} & 
        \textbf{0.783} &
        \underline{\textbf{1.758}} &
        \underline{\textbf{3.896}} & 
        \underline{\textbf{1.690}} & 
        \underline{\textbf{2.175}} \\
        \bottomrule
        \end{tabular}
    }
    \label{tab:generalizability}
\end{table}

%% file: tables/alpha_scaling.tex
\begin{table}[H]
\centering
\caption{Cross-market average performance across different numbers of $N_\alpha$ under our inter-stock design (b), loss design (c), and capacity scaling (d), as defined in \cref{tab:component_ablation}. Multi-seed Ensemble trains $N_\alpha$ independent instances of configuration (b) separately and averages predictions. Bold indicates configuration (c) or (d) outperforms Multi-seed Ensemble on the corresponding cell. \underline{Underline} denotes the best among all methods.}
\label{tab:alpha_scaling}
\resizebox{\columnwidth}{!}{%
\begin{tabular}{lccc|ccc|ccc}
\toprule
 & \multicolumn{3}{c}{\textbf{Multi-seed Ensemble}} & \multicolumn{6}{c}{\textbf{Single-model, $N_\alpha$-preds}} \\
\cmidrule(lr){2-4} \cmidrule(lr){5-10}
{\textbf{Config}} & \multicolumn{3}{c}{\textbf{Inter-Stock Design (b)}} & \multicolumn{3}{c}{\textbf{+ Loss Design (c)}} & \multicolumn{3}{c}{\textbf{+ Capacity Scaling (d)}} \\
\textbf{$N_\alpha$} & \textbf{AR} & \textbf{SR} & \textbf{CR} & \textbf{AR} & \textbf{SR} & \textbf{CR} & \textbf{AR} & \textbf{SR} & \textbf{CR} \\
\midrule
1  & 0.684 & 1.430 & 1.749 & \textbf{1.067} & \textbf{1.526} & \textbf{1.791} & \textbf{1.196} & \textbf{1.566} & \textbf{1.929} \\
2  & 0.679 & 1.528 & 1.886 & \textbf{0.916} & \textbf{1.531} & 1.832 & \textbf{1.099} & \textbf{1.593} & \textbf{2.020} \\
4  & 0.656 & 1.562 & 1.989 & \textbf{0.834} & 1.532 & 1.925 & \textbf{1.083} & \textbf{1.653} & \textbf{2.107} \\
8  & 0.631 & 1.615 & 2.076 & \textbf{0.830} & 1.579 & 1.996 & \textbf{1.208} & \textbf{1.660} & 2.031 \\
16 & 0.636 & 1.659 & 2.134 & \textbf{0.847} & 1.530 & 1.799 & \textbf{1.225} & 1.653 & 2.095 \\
24 & 0.637 & 1.665 & 2.117 & \textbf{0.886} & 1.532 & 1.793 & \underline{\textbf{1.297}} & \underline{\textbf{1.690}} & \underline{\textbf{2.175}} \\
\bottomrule
\end{tabular}%
}
\end{table}

%% file: tables/diversity_sensitivity.tex
\begin{table}[H]
\centering
\caption{Cross-market average alpha correlation and performance across different values of the diversity weight $\lambda$ under our loss design (c), capacity scaling (d, $N_\alpha=8$), and alpha scaling (e, $N_\alpha=24$), as defined in \cref{tab:component_ablation}. Bold and \underline{underline} indicate the peak and second-highest value in each column.}
\label{tab:diversity_sensitivity}
\resizebox{\columnwidth}{!}{%
\begin{tabular}{lcccc|cccc|cccc}
\toprule
\multirow{2}{*}{\textbf{Config}} & \multicolumn{4}{c}{\textbf{Loss Design (c)}} & \multicolumn{4}{c}{\textbf{+ Capacity Scaling (d)}} & \multicolumn{4}{c}{\textbf{+ Alpha Scaling (e)}} \\
  & \multicolumn{4}{c}{\textbf{$N_\alpha=8$}} & \multicolumn{4}{c}{\textbf{$N_\alpha=8$}} & \multicolumn{4}{c}{\textbf{$N_\alpha=24$}} \\
 $\boldsymbol{\lambda}$ & \textbf{Corr} & \textbf{AR} & \textbf{SR} & \textbf{CR} & \textbf{Corr} & \textbf{AR} & \textbf{SR} & \textbf{CR} & \textbf{Corr} & \textbf{AR} & \textbf{SR} & \textbf{CR} \\
\midrule
0.00 & \textbf{0.964} & \textbf{1.035} & 1.510 & 1.856 & \textbf{0.966} & \textbf{1.333} & 1.563 & 1.891 & \textbf{0.967} & \underline{1.300} & 1.526 & 1.804 \\
0.05 & \underline{0.885} & \underline{0.922} & 1.512 & 1.754 & \underline{0.885} & \underline{1.292} & 1.604 & 1.979 & \underline{0.910} & \textbf{1.307} & 1.586 & 1.927 \\
0.10 & 0.671 & 0.830 & \textbf{1.579} & \textbf{1.996} & 0.672 & 1.208 & \underline{1.660} & 2.031 & 0.731 & 1.297 & \textbf{1.690} & \textbf{2.175} \\
0.15 & 0.471 & 0.715 & \underline{1.514} & \underline{1.858} & 0.496 & 0.951 & \textbf{1.683} & \textbf{2.147} & 0.540 & 0.971 & \underline{1.664} & \textbf{2.175} \\
0.20 & 0.362 & 0.640 & 1.504 & 1.874 & 0.375 & 0.772 & 1.633 & \underline{2.047} & 0.437 & 0.819 & 1.640 & \underline{2.111} \\
0.30 & 0.315 & 0.590 & 1.496 & 1.847 & 0.301 & 0.640 & 1.568 & 1.928 & 0.314 & 0.608 & 1.562 & 1.949 \\
0.40 & 0.366 & 0.591 & 1.500 & 1.824 & 0.324 & 0.639 & 1.554 & 1.829 & 0.254 & 0.496 & 1.476 & 1.767 \\
\bottomrule
\end{tabular}%
}
\end{table}